
\def\ignore#1{}
 

\newcount\sectnum
\newcount\subsectnum
\newcount\eqnumber

\global\eqnumber=1\sectnum=0


\def\lab{(\the\sectnum.\the\eqnumber)}



\def\show#1{#1}



\def\smskip{\vskip 5 pt}
\def\medskip{\vskip 10 pt}
\def\bigskip{\vskip 15 pt}
\def\pn{\par\noindent}
\def\br{\break}

\def\bl{\bigl} 
\def\br{\bigr} 
\def\lf{\left}
\def\ri{\right}

\def\frac#1#2{{#1\over #2}}

\def\ol#1{\overline{#1}}

\def\a{\alpha}

\def\l{\lambda}
\def\g{\gamma}
\def\m{\mu}

\def\e{\epsilon}

\def\re{\Re}
\def\rn{\Re^n}

\def\tl{\tilde}

\def\old#1{}
\def\leaderfill{\leaders\hbox to 1em{\hss.\hss}\hfill}


\parindent=2pc
\baselineskip=15pt
\vsize=8.7 true in
\voffset=0.125 true in
\parskip=3pt


\def\minprob#1#2#3{$$\eqalign{&\hbox{minimize\ \ }#1\cr &\hbox{subject to\ \
}#2\cr}\ifnum 0=#3{}\else\eqno(#3)\fi$$}        
     
\def\maxprob#1#2#3{$$\eqalign{&\hbox{maximize\ \ }#1\cr &\hbox{subject to\ \
}#2\cr}\ifnum 0=#3{}\else\eqno(#3)\fi$$}        
     
\def\aligntwo#1#2#3#4#5{$$\eqalign{#1&#2\cr #3&#4\cr}
\ifnum 0=#5{}\else\eqno(#5)\fi$$}
\def\alignthree#1#2#3#4#5#6#7{$$\eqalign{#1&#2\cr #3&#4\cr #5&#6\cr}
\ifnum 0=#7{}\else\eqno(#7)\fi$$}


\def\eqnum{\eqno{\hbox{(\the\sectnum.\the\eqnumber)}\global\advance\eqnumber
by1}}

\def\eqnu{\eqno{\hbox{(\the\sectnum.\the\eqnumber)}\global\advance\eqnumber
by1}}

\newcount\examplnumber
\def\examplnum{\global\advance\examplnumber by1}

\newcount\figrnumber
\def\figrnum{\global\advance\figrnumber by1}

\newcount\propnumber
\def\propnum{\global\advance\propnumber by1}

\newcount\defnumber
\def\defnum{\global\advance\defnumber by1}

\newcount\lemmanumber
\def\lemmanum{\global\advance\lemmanumber by1}

\newcount\assumptionnumber
\def\assumptionnum{\global\advance\assumptionnumber by1}

\newcount\conditionnumber
\def\conditionnum{\global\advance\conditionnumber by1}

\def\exampl{\the\sectnum.\the\examplnumber}
\def\figr{\the\sectnum.\the\figrnumber}
\def\propn{\the\sectnum.\the\propnumber}
\def\defn{\the\sectnum.\the\defnumber}
\def\lemman{\the\sectnum.\the\lemmanumber}
\def\assumptionn{\the\sectnum.\the\assumptionnumber}
\def\condn{\the\sectnum.\the\conditionnumber}

\def\section#1{\goodbreak\vskip 3pc plus 6pt minus 3pt\leftskip=-2pc
   \global\advance\sectnum by 1\eqnumber=1
\global\examplnumber=1\figrnumber=1\propnumber=1\defnumber=1\lemmanumber=1\assumptionnumber=1 \conditionnumber =1%
   \line{\hfuzz=1pc{\hbox to 3pc{\bf 
   \vtop{\hfuzz=1pc\hsize=38pc\hyphenpenalty=10000\noindent\uppercase{\the\sectnum.\quad #1}}\hss}}
			\hfill}
			\leftskip=0pc\nobreak\tenf
			\vskip 1pc plus 4pt minus 2pt\noindent\ignorespaces}



\def\sect#1{\noindent\leftskip=-2pc\tenf
   \goodbreak\vskip 1pc plus 4pt minus 2pt
                \global\advance\subsectnum by 1\eqnumber=1
   \line{\hfuzz=1pc{\hbox to 3pc{\bf 
   \vtop{\hfuzz=1pc\hsize=38pc\hyphenpenalty=10000\noindent\uppercase{{\bf #1}}}\hss}}
                        \hfill}
   \leftskip=0pc\nobreak\tenf
                        \vskip 1pc plus 4pt minus 2pt\nobreak\noindent\ignorespaces}

\def\subsection#1{\noindent\leftskip=0pc\tenf
   \goodbreak\vskip 1pc plus 4pt minus 2pt
   \line{\hfuzz=1pc{\hbox to 3pc{\bf 
   \vtop{\hfuzz=1pc\hsize=38pc\hyphenpenalty=10000\noindent{\bf #1}}\hss}}
                        \hfill}
   \leftskip=0pc\nobreak\tenf
                        \vskip 1pc plus 4pt minus 2pt\nobreak\noindent\ignorespaces}
\def\subsubsection#1{\goodbreak\vskip 1pc plus 4pt minus 2pt
   \hfuzz=3pc\leftskip=0pc\noindent\tenit #1 \nobreak\tenf\vskip 6pt plus 1pt
                                minus 1pt\nobreak\ignorespaces\leftskip=0pc}
%

\def\beginexample#1{\noindent\goodbreak\vskip 6pt plus 1pt minus 1pt
\noindent
  \hbox {\bf Example #1\hss}
  \nobreak\vskip 4pt plus 1pt minus 1pt \nobreak\noindent\ninef
  \global\advance
                \leftskip by\parindent\pn}
\def\endexample{\vskip 12pt\tenf\par
  \global\advance\leftskip by -\parindent
  }

\def\beginexercise#1{\noindent\goodbreak\vskip 6pt plus 1pt minus 1pt \noindent\global\normalbaselineskip=12pt
  \hbox {\bf Exercise #1\hss}
  \nobreak\vskip 4pt plus 1pt minus 1pt 
  \nobreak\noindent\ninef\global\advance\leftskip
                        by\parindent\pn}
\def\endexercise{\vskip 12pt\tenf\par
  \global\advance\leftskip by -\parindent
  }

\def\beginsection#1{\noindent\goodbreak\vskip 6pt plus 1pt minus 1pt \noindent\global\normalbaselineskip=12pt
  \hbox {\it #1\hss}
  \vskip 0.1pt plus 1pt minus 1pt \nobreak\noindent\ninef\global\advance
                \leftskip by\parindent\noindent\pn}
\def\endsection{\vskip 12pt\tenf\par
  \global\advance\leftskip by -\parindent
}

%


\def\proposition#1{\smskip\pn{\bf Proposition #1}\quad}
\def\proof{\smskip\pn{\bf Proof:}\quad}

 \def\qed{\quad{\bf
Q.E.D.} \par\bigskip}
\def\ref{\smskip\pn}

\def\chapter#1#2{{\bf \centerline{\helbigbig
{#1}}}\bigskip\bigskip{\bf \centerline{\helbigbig
{#2}}}\bigskip\bigskip} 



\def\longpapertitle#1#2#3{{\bf \centerline{\helbigb
{#1}}}\bigskip{\bf \centerline{\helbigb
{#2}}}\bigskip\bigskip{\centerline{
by}}\bigskip{\bf \centerline{
{#3}}}\bigskip\bigskip} 


\def\nitem#1{\smskip\item{#1}}

\newcount\alphanum
\newcount\romnum

\def\alphaenumerate{\ifcase\alphanum \or (a)\or (b)\or (c)\or (d)\or (e)\or
(f)\or (g)\or (h)\or (i)\or (j)\or (k)\fi}
\def\romenumerate{\ifcase\romnum \or (i)\or (ii)\or (iii)\or (iv)\or (v)\or
(vi)\or (vii)\or (viii)\or (ix)\or (x)\or (xi)\fi}

\def\alist{\begingroup\vskip10pt\alphanum=1
\parskip=2pt\parindent=0pt \leftskip=3pc
\everypar{\llap{{\rm\alphaenumerate\hskip1em}}\advance\alphanum by1}}

\def\nolist{\begingroup\vskip10pt\alphanum=0
\parskip=2pt\parindent=0pt \leftskip=3pc
\everypar{\llap{\global\advance\alphanum by1(\the\alphanum)\hskip1em}}}

\def\romlist{\begingroup\vskip10pt\romnum=1
\parskip=2pt\parindent=0pt \leftskip=5pc
\everypar{\llap{{\rm\romenumerate\hskip1em}}\advance\romnum by1}}



\long\def\fig#1#2#3{\vbox{\vskip1pc\vskip#1
\prevdepth=12pt \baselineskip=12pt
\vskip1pc
\hbox to\hsize{\hfill\vtop{\hsize=25pc\noindent{\eightbf Figure #2\ }
{\eightpoint#3}}\hfill}}}

\long\def\widefig#1#2#3{\vbox{\vskip1pc\vskip#1
\prevdepth=12pt \baselineskip=12pt
\vskip1pc
\hbox to\hsize{\hfill\vtop{\hsize=28pc\noindent{\eightbf Figure #2\ }
{\eightpoint#3}}\hfill}}}

\long\def\table#1#2{\vbox{\vskip0.5pc
\prevdepth=12pt \baselineskip=12pt
\hbox to\hsize{\hfill\vtop{\hsize=25pc\noindent{\eightbf Table #1\ }
{\eightpoint#2}}\hfill}}}

 
\def\rightheadline#1{\headline{\tenrm\hfil #1}}


\long\def\leftfig#1#2{\vbox{\smskip\hsize=220pt
\vtop{{\noindent {\bf #1}}}
\smskip
\noindent
\vbox{{\noindent #2}}
}}

\long\def\rightfig#1#2#3{\vbox{\smskip\vskip#1
\prevdepth=12pt \baselineskip=12pt
\hsize=210pt
\smskip
\vbox{\noindent{\eightbold #2}
\hskip1em{\eightpoint#3}}
}}

\long\def\concept#1#2#3#4#5{\bigskip\hrule
\vbox{\hbox{\leftfig{#1}{#2} \hskip3em
\rightfig{#3}{#4}{#5}} \smskip}
\hrule\bigskip}


\long\def\bconcept#1#2#3#4#5#6#7{
\vbox{
\hbox to \hsize{\vtop{\par #1}}
\concept{#2}{#3}{#4}{#5}{#6}
\hbox to \hsize{\vtop{\par #7}}
\smskip}
}




\def\boxit#1{\vbox{\hrule\hbox{\vrule\kern3pt
                                \vbox{\kern3pt#1\kern3pt}\kern3pt\vrule}\hrule}}
\def\centerboxit#1{$$\vbox{\hrule\hbox{\vrule\kern3pt
                                \vbox{\kern3pt#1\kern3pt}\kern3pt\vrule}\hrule}$$}

\long\def\boxtext#1#2{$$\boxit{\vbox{\hsize #1\noindent\strut #2\strut}}$$}

%
%
%

\def\picture #1 by #2 (#3){
  \vbox to #2{
    \hrule width #1 height 0pt depth 0pt
    \vfill
    \special{picture #3} 
    }
  }

\def\scaledpicture #1 by #2 (#3 scaled #4){{
  \dimen0=#1 \dimen1=#2
  \divide\dimen0 by 1000 \multiply\dimen0 by #4
  \divide\dimen1 by 1000 \multiply\dimen1 by #4
  \picture \dimen0 by \dimen1 (#3 scaled #4)}
  }

%
%

\long\def\captfig#1#2#3#4#5{\vbox{\vskip1pc
\hbox to\hsize{\hfill{\picture #1 by #2 (#3)}\hfill}
\prevdepth=9pt \baselineskip=9pt
\vskip1pc
\hbox to\hsize{\hfill\vtop{\hsize=24pc\noindent{\eightbold Figure #4}
\hskip1em{\eightpoint#5}}\hfill}}}

%
%
%

\def\illustration #1 by #2 (#3){
  \vskip#2\hskip#1\special{illustration #3} 
    }

\def\scaledillustration #1 by #2 (#3 scaled #4){{
  \dimen0=#1 \dimen1=#2
  \divide\dimen0 by 1000 \multiply\dimen0 by #4
  \divide\dimen1 by 1000 \multiply\dimen1 by #4
  \illustration \dimen0 by \dimen1 (#3 scaled #4)}
  }


\newbox\graybox
\newdimen\xgrayspace
\newdimen\ygrayspace
%
%
%
%
%
%
%
%
%

\def\Textshade#1#2#3#4#5#6{%
    \xgrayspace=#4pt%
    \ygrayspace=#4pt%
    \def\grayshade{#3}%
    \def\linewidth{#5}%
    \def\theradius{#6}%
    \setbox\graybox=\hbox{\surroundboxa{#2}}%
    \hbox{%
    \hbox to 0pt{%
    \PScommands
}%
    \box\graybox}}%
%
%
\long%

\long%
\def\Parashade#1#2#3#4#5#6#7{%
    \xgrayspace=#4pt%
    \ygrayspace=#4pt%
    \def\grayshade{#3}%
    \def\linewidth{#5}%
    \def\theradius{#6}%
    \def\thevskip{#7pt}%
    \setbox\graybox=\hbox{\surroundboxb{#2}}%
    \vskip\thevskip%
    \hbox{%
    \hbox to 0pt{%
    \PScommands
}%
     \box\graybox}%
     \vskip\thevskip%
}%
%
%
%
\long\def\surroundboxa#1{\leavevmode\hbox{\vtop{%
\vbox{\kern\ygrayspace%
\hbox{\kern\xgrayspace#1%
      \kern\xgrayspace}}\kern\ygrayspace}}}
%
%
\long\def\surroundboxb#1{\leavevmode\hbox{\vtop{%
\vbox{\kern\ygrayspace%
\hbox{\kern\xgrayspace\vbox{\advance\hsize-2\xgrayspace#1}%
      \kern\xgrayspace}}\kern\ygrayspace}}}
%
%
%
\long\def\PScommands{%
\special{rawpostscript
/sharpbox{%
           newpath
           xmin ymin moveto
           xmin ymax lineto
           xmax ymax lineto
           xmax ymin lineto
           xmin ymin lineto
           closepath 
          }bind def
}%
\special{rawpostscript
/sharpboxnb{%
           newpath
           xmin ymin moveto
           xmin ymax lineto
           xmax ymax lineto
           xmax ymin lineto
          }bind def
}%
\special{rawpostscript
/sharpboxnt{%
           newpath
           xmin ymax moveto
           xmin ymin lineto
           xmax ymin lineto
           xmax ymax lineto
          }bind def
}%
\special{rawpostscript
/roundbox{%
           newpath
           xmin radius add ymin moveto
           xmax ymin xmax ymax radius arcto
           xmax ymax xmin ymax radius arcto
           xmin ymax xmin ymin radius arcto
           xmin ymin xmax ymin radius arcto 16 {pop} repeat
           closepath
          }bind def
}%
\special{rawpostscript
/sharpcorners{%
               sharpbox gsave grayshade setgray fill grestore 
               linewidth setlinewidth stroke
              }bind def
}%
\special{rawpostscript
/sharpcornersnt{%
               sharpboxnt gsave grayshade setgray fill grestore 
               linewidth setlinewidth stroke
              }bind def
}%
\special{rawpostscript
/sharpcornersnb{%
               sharpboxnb gsave grayshade setgray fill grestore 
               linewidth setlinewidth stroke
              }bind def
}%
\special{rawpostscript
/roundcorners{%
               roundbox gsave grayshade setgray fill grestore 
               linewidth setlinewidth stroke
              }bind def
}%
\special{rawpostscript
/plainbox{%
           sharpbox grayshade setgray fill 
          }bind def
}%
%
\special{rawpostscript
/roundnoframe{%
               roundbox grayshade setgray fill 
              }bind def
}%
\special{rawpostscript
/sharpnoframe{%
               sharpbox grayshade setgray fill 
              }bind def
}%
}%
%
%

\def\pshade#1{\Parashade{sharpcorners}{#1}{0.95}{10}{0.5}{10}{10}}


\def\boxit#1{\vbox{\hrule\hbox{\vrule\kern3pt
                                \vbox{\kern3pt#1\kern3pt}\kern3pt\vrule}\hrule}}

\def\boxitnb#1{\vbox{\hrule\hbox{\vrule\kern3pt
                                \vbox{\kern3pt#1\kern3pt}\kern3pt\vrule}}}

\def\boxitnt#1{\vbox{\hbox{\vrule\kern3pt
                                \vbox{\kern3pt#1\kern3pt}\kern3pt\vrule}\hrule}}

\long\def\boxtext#1#2{$$\boxit{\vbox{\hsize #1\noindent\strut #2\strut}}$$}



\def\texshopbox#1{\boxtext{462pt}{\vskip-1.5pc\pshade{\vskip-1.0pc#1\vskip-2.0pc}}}


%
%
%
%
%
%
%
%
\font\helbigbig=cmr10 scaled 2500%
\font\helbigb=cmbx10 scaled 1500%
\font\eightbold=cmbx8%

\def\tenf{\hel}%
\def\tenit{\heli}%
\def\ninef{\ninehel}%
\def\nineit{\nineheli}%
%
%


\font\tenrm=cmr10%
\font\teni=cmmi10%
\font\tensy=cmsy10%
\font\tenbf=cmbx10%
\font\tentt=cmtt10%
\font\tenit=cmti10%
\font\tensl=cmsl10%

\def\tenpoint{\def\rm{\fam0\tenrm}%
\textfont0=\tenrm%
\textfont1=\teni%
\textfont2=\tensy%
\textfont\itfam=\tenit%
\textfont\slfam=\tensl%
\textfont\ttfam=\tentt%
\textfont\bffam=\tenbf%
\scriptfont0=\sevenrm%
\scriptfont1=\seveni%
\scriptfont2=\sevensy%
\scriptscriptfont0=\sixrm%
\scriptscriptfont1=\sixi%
\scriptscriptfont2=\sixsy%
\def\it{\fam\itfam\tenit}%
\def\tt{\fam\ttfam\tentt}%
\def\sl{\fam\slfam\tensl}%
\scriptfont\bffam=\sevenbf%
\scriptscriptfont\bffam=\sixbf%
\def\bf{\fam\bffam\tenbf}%
\normalbaselineskip=18pt%
\normalbaselines\rm}%

\font\ninerm=cmr9%
\font\ninebf=cmbx9%
\font\nineit=cmti9%
\font\ninesy=cmsy9%
\font\ninei=cmmi9%
\font\ninett=cmtt9%
\font\ninesl=cmsl9%

\def\ninepoint{\def\rm{\fam0\ninerm}%
\textfont0=\ninerm%
\textfont1=\ninei%
\textfont2=\ninesy%
\textfont\itfam=\nineit%
\textfont\slfam=\ninesl%
\textfont\ttfam=\ninett%
\textfont\bffam=\ninebf%
\scriptfont0=\sixrm%
\scriptfont1=\sixi%
\scriptfont2=\sixsy%
\def\it{\fam\itfam\nineit}%
\def\tt{\fam\ttfam\ninett}%
\def\sl{\fam\slfam\ninesl}%
\scriptfont\bffam=\sixbf%
\scriptscriptfont\bffam=\fivebf%
\def\bf{\fam\bffam\ninebf}%
\normalbaselineskip=16pt%
\normalbaselines\rm}%

\font\eightrm=cmr8%
\font\eighti=cmmi8%
\font\eightsy=cmsy8%
\font\eightbf=cmbx8%
\font\eighttt=cmtt8%
\font\eightit=cmti8%
\font\eightsl=cmsl8%

\def\eightpoint{\def\rm{\fam0\eightrm}%
\textfont0=\eightrm%
\textfont1=\eighti%
\textfont2=\eightsy%
\textfont\itfam=\eightit%
\textfont\slfam=\eightsl%
\textfont\ttfam=\eighttt%
\textfont\bffam=\eightbf%
\scriptfont0=\sixrm%
\scriptfont1=\sixi%
\scriptfont2=\sixsy%
\scriptscriptfont0=\fiverm%
\scriptscriptfont1=\fivei%
\scriptscriptfont2=\fivesy%
\def\it{\fam\itfam\eightit}%
\def\tt{\fam\ttfam\eighttt}%
\def\sl{\fam\slfam\eightsl}%
\scriptscriptfont\bffam=\fivebf%
\def\bf{\fam\bffam\eightbf}%
\normalbaselineskip=14pt%
\normalbaselines\rm}%

\font\sevenrm=cmr7%
\font\seveni=cmmi7%
\font\sevensy=cmsy7%
\font\sevenbf=cmbx7%

\def\sevenpoint{%
   \def\rm{\sevenrm}\def\bf{\sevenbf}%
   \def\smc{\sevensmc}\baselineskip=12pt\rm}%

\font\sixrm=cmr6%
\font\sixi=cmmi6%
\font\sixsy=cmsy6%
\font\sixbf=cmbx6%

\fontdimen13\tensy=2.6pt%
\fontdimen14\tensy=2.6pt%
\fontdimen15\tensy=2.6pt%
\fontdimen16\tensy=1.2pt%
\fontdimen17\tensy=1.2pt%
\fontdimen18\tensy=1.2pt%

\def\tenf{\tenpoint}%
\def\ninef{\ninepoint}%
%



\def\texshopbox#1{\boxtext{462pt}{\vskip-1.5pc\pshade{\vskip-1.0pc#1\vskip-2.0pc}}}


\input miniltx

\ifx\pdfoutput\undefined
  \def\Gin@driver{dvips.def} 
\else
  \def\Gin@driver{pdftex.def} 
\fi

\input graphicx.sty
\resetatcatcode

\long\def\fig#1#2#3{\vbox{\vskip1pc\vskip#1
\prevdepth=12pt \baselineskip=12pt
\vskip1pc
\hbox to\hsize{\hfill\vtop{\hsize=30pc\noindent{\eightbf Figure #2\ }
{\eightpoint#3}}\hfill}}}

\def\show#1{}



\def\section#1{\goodbreak\vskip 3pc plus 6pt minus 3pt\leftskip=-2pc
   \global\advance\sectnum by 1\eqnumber=1
\global\examplnumber=1\figrnumber=1\propnumber=1\defnumber=1\lemmanumber=1\assumptionnumber=1\subsectnum=0%
   \line{\hfuzz=1pc{\hbox to 3pc{\bf 
   \vtop{\hfuzz=1pc\hsize=38pc\hyphenpenalty=10000\noindent\uppercase{\the\sectnum.\quad #1}}\hss}}
			\hfill}
			\leftskip=0pc\nobreak\tenf
			\vskip 1pc plus 4pt minus 2pt\noindent\ignorespaces}

\def\subsection#1{\noindent\leftskip=0pc\tenf
   \goodbreak\vskip 1pc plus 4pt minus 2pt
               \global\advance\subsectnum by 1
   \line{\hfuzz=1pc{\hbox to 3pc{\bf \the\sectnum.\the\subsectnum.\ \ \
   \vtop{\hfuzz=1pc\hsize=38pc\hyphenpenalty=10000\noindent{\bf #1}}\hss}}
                        \hfill}
   \leftskip=0pc\nobreak\tenf
                        \vskip 1pc plus 4pt minus 2pt\nobreak\noindent\ignorespaces}

\def\subsubsection#1{\goodbreak\vskip 1pc plus 4pt minus 2pt
   \hfuzz=3pc\leftskip=0pc\noindent{\bf #1} \nobreak\vskip 6pt plus 1pt
                                minus 1pt\nobreak\ignorespaces\leftskip=0pc}


\rightheadline{\botmark}

\pageno=1

\def\longpapertitle#1#2#3{{\bf \centerline{\helbigb
{#1}}}\medskip{\bf \centeline{\helbigb
{#2}}}\bigskip{\bf \centerline{
{#3}}}\bigskip}

\vskip-3pc

\def\xstar{X^{\raise0.04pt\hbox{\sevenpoint *}} }

\def\jstar{J^{\raise0.04pt\hbox{\sevenpoint *}} }

\rightheadline{\botmark}

\pageno=1

\rightheadline{\botmark}

\pageno=1

\rightheadline{\botmark}

\pn {\bf October 2019}
\bigskip \bigskip\bigskip

\bigskip

\def\longpapertitle#1#2#3{{\bf \centerline{\helbigb
{#1}}}\medskip{\bf \centerline{\helbigb
{#2}}}\bigskip{\bf \centerline{
{#3}}}\bigskip}

\vskip-3pc

\longpapertitle{Biased Aggregation, Rollout, and}{Enhanced Policy Improvement for Reinforcement Learning\footnote{\dag}
{\ninepoint  An abbreviated version of this paper was included as Section 6.5 in the author's book ``Reinforcement Learning and Optimal Control;" [Ber19].}}{ {Dimitri P.\ Bertsekas\footnote{\ddag}
{\ninepoint  The author is with the Dept.\ of Electr.\ Engineering and
Comp.\ Science, and the Laboratory for Information and Decision Systems (LIDS), M.I.T., Cambridge, Mass., 02139. Thanks are due to John Tsitsiklis for helpful comments.}}}

We propose a new aggregation framework for approximate dynamic programming, which provides a connection with rollout algorithms, approximate policy iteration, and other single and multistep lookahead methods. The central novel characteristic is the use of a {\it bias function} $V$ of the state, which biases the values of the aggregate cost function towards their correct levels. 
The classical aggregation framework is obtained when $V\equiv0$, but our scheme works best when $V$ is a known reasonably good approximation to the optimal cost function $\jstar$.
 
When $V$ is equal to the cost function $J_\m$ of some known policy $\m$ and there is only one aggregate state, our scheme is equivalent to the rollout algorithm based on $\m$ (i.e., the result of a single policy improvement starting with the policy $\m$). When $V=J_\m$ and there are multiple aggregate states, our aggregation approach can be used as a more powerful form of improvement of $\m$. Thus, when combined with an approximate policy evaluation scheme, our approach can form the basis for a new and enhanced form of approximate policy iteration.

When $V$ is  a generic bias function, our scheme is equivalent to approximation in value space with lookahead function equal to $V$ plus a local correction within each aggregate state. The local correction levels are obtained by solving a low-dimensional aggregate DP problem,  yielding an arbitrarily close approximation to $\jstar$, when the number of aggregate states is sufficiently large. 
Except for the bias function, the aggregate DP problem is similar to the one of the classical aggregation framework, and its algorithmic solution by simulation or other methods is nearly identical to one for classical aggregation, assuming values of $V$ are available when needed. 

\vskip-1.5pc
\vfill\eject

\section{Introduction}

\vskip-0.5pc

\pn We introduce an extension of the classical aggregation framework for approximate dynamic programming (DP for short). We will focus on the standard discounted infinite horizon problem with state space  
${\cal S}=\{1,\ldots,n\},$  although the ideas apply more broadly.
State transitions $(i,j)$, $i,j\in{\cal S}$, under control $u$ occur at discrete times according to transition probabilities $p_{ij}(u)$, and generate a cost $\a^k g(i,u,j)$ at time $k$, where $\a\in(0,1)$ is the discount factor. We consider deterministic stationary policies $\m$ such that for each $i$, $\m(i)$ is a control that belongs to a finite constraint set $U(i)$. We denote by $J_\m(i)$ the total discounted expected cost  of $\m$ over an infinite number of stages starting from state $i$, and by $\jstar(i)$ the minimal value of $J_\m(i)$ over all $\m$. We denote by $J_\m$ and $\jstar $ the $n$-dimensional vectors that have components $J_\m(i)$ and $\jstar(i)$, $i\in{\cal S}$, respectively. As is well known, $J_\m$ is the unique solution of the Bellman equation for policy $\m$:
 $$J_\m(i) = \sum_{j=1}^n
p_{ij}\bl(\mu(i)\br)\Big(g\bl(i,\mu(i),j\br) + \a J_\m(j)\Big),\qquad  i\in{\cal S},\eqnum\show{oneo}$$
while $\jstar $ is the unique solution of the Bellman equation
$$\jstar (i) = \min_{u\in U(i)}\sum_{j=1}^n p_{ij}(u)\big(g(i,u,j) + \a \jstar (j)\big),
\qquad i\in{\cal S}.\xdef\onef{\lab}\eqnum\show{oneo}$$

\vskip-0.5pc

Aggregation can be viewed as a problem approximation approach: we approximate the original problem with a related ``aggregate" problem, which we  solve exactly by some form of DP. We thus obtain a policy that is optimal for the aggregate problem but is suboptimal for the original. In the present paper we will extend this framework so that it can be used more flexibly and more broadly. The key idea is to enhance aggregation with prior knowledge, which may be obtained using any approximate DP technique, possibly unrelated to aggregation.

Our point of departure is the classical aggregation framework that is discussed in the author's two-volume textbook [Ber12], [Ber17] (Section 6.2.3 of Vol.\ I, and Sections 6.1.3 and 6.4 of Vol.\ II). This framework is itself a formalization of earlier aggregation ideas, which originated in scientific computation and operations research (see for example Bean, Birge, and Smith [BBS87], Chatelin and Miranker [ChM82], Douglas and Douglas [DoD93], Mendelssohn [Men82], Rogers et.\ al.\ [RPW91], and Vakhutinsky, Dudkin, and Ryvkin [VDR79]). Common examples of aggregation within this context arise in discretization of continuous spaces, and approximation of fine grids by coarser grids, as in multigrid methods. Aggregation was introduced in the simulation-based approximate DP context, mostly in the form of value iteration; see Singh, Jaakkola, and Jordan [SJJ95], Gordon [Gor95], Tsitsiklis and Van Roy [TsV96] (also the book by Bertsekas and Tsitsiklis [BeT96], Sections 3.1.2 and 6.7).  More recently, aggregation was discussed in the context of partially observed Markovian decision problems (POMDP) by Yu and Bertsekas [YuB12], and in 
a reinforcement learning context involving the notion of ``options" by Ciosek and Silver [CiS15], and the notion of ``bottleneck simulator" by Serban et.\ al.\ [SSP18]; in all these cases encouraging computational results were presented. 

In a common type of classical aggregation, we group the states $1,\ldots,n$ into disjoint subsets, viewing each subset as a state of an ``aggregate DP problem." We then solve this problem exactly, to obtain a piecewise constant approximation to the optimal cost function $\jstar$: the approximation is constant  within each subset, at a level that is an ``average" of the values of $\jstar$ within the subset. This is called  {\it hard aggregation\/}. The state space partition in hard aggregation is arbitrary, but is often determined by using a vector of ``features" of the state (states with ``similar" features are grouped together). Features can generally form the basis for specifying aggregation architectures, as has been discussed in Section 3.1.2 of the neuro-dynamic programming book [BeT96], in Section 6.5 of the author's DP book [Ber12], and in the recent survey [Ber18a].  While feature-based selection of the aggregate states can play an important role within our framework, it will not be discussed much, because it is not central to the ideas of this paper.

In fact there are many different ways to form the aggregate states and to specify the aggregate DP problem.  We will not go very much into the details of specific schemes in this paper, referring to earlier literature for further discussion. We just mention {\it aggregation with representative states\/}, where the starting point is a finite collection of states, which we view as ``representative."  The representative states may also be selected using features (see [Ber18a]).
\xdef\figbiasedaggr{\figr}\figrnum\show{myfigure}

Our proposed extended framework, which we call {\it biased aggregation\/}, is similar in structure to classical aggregation, but involves in addition a function/vector
 $$V=\big(V(1),\ldots,V(n)\big)$$
 of the state, called the {\it bias function\/}, which affects the cost structure of the aggregate problem, and  biases the values of its optimal cost function towards their correct levels. With $V=0$, we obtain the classical aggregation scheme. With $V\ne0$, biased aggregation yields an approximation to $\jstar$ that is equal to $V$ plus a local correction; see Fig.\ \figbiasedaggr. In hard aggregation the local correction is piecewise constant (it is constant within each aggregate state). In other forms of aggregation it is piecewise linear. Thus the aggregate DP problem provides a correction to $V$, which may itself be a reasonably good estimate of $\jstar$.

\topinsert{
\centerline{\hskip0pc\includegraphics[width=5in]{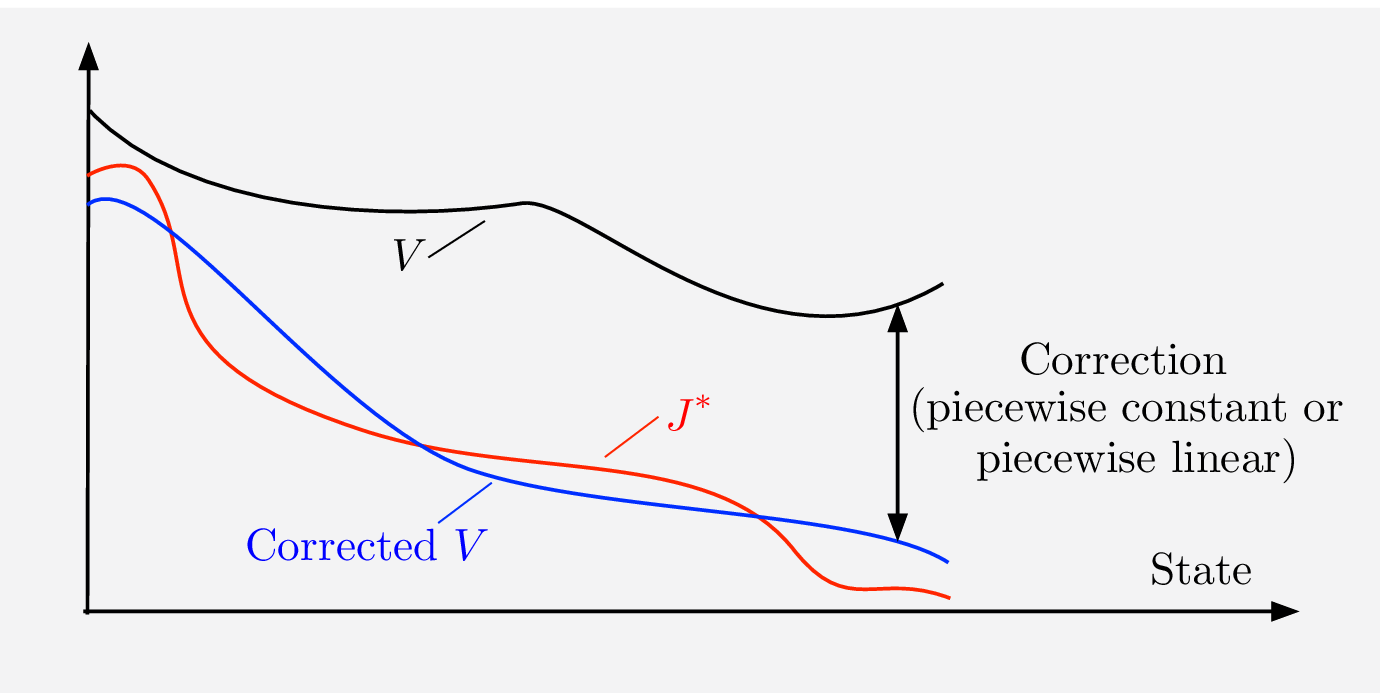}}
\fig{-0.5pc}{\figbiasedaggr}{Schematic illustration of biased aggregation. It provides an approximation to $J^*$ that is equal to  the bias function $V$ plus a local correction. When $V=0$, we obtain the classical aggregation framework.\smskip}
}
\endinsert

An obvious context where biased aggregation can be used is to improve on an approximation to $\jstar$ obtained using a different method, such as for example by neural network-based approximate policy iteration, by rollout, or by exact DP applied to a simpler version of the given problem. Another context, is to integrate the choice of $V$ with the choice of the aggregation framework with the aim of improving on classical aggregation. Generally, if $V$ captures a fair amount of the nonlinearity of $\jstar$, we speculate that a major benefit of the bias function approach will be a reduction of the number of aggregate states needed for adequate performance, and the attendant facilitation of the algorithmic solution of the aggregate problem. 
 
The bias function idea is new to our knowledge within the context of aggregation.
The closest connection to the existing aggregation literature is the adaptive aggregation framework of the paper by Bertsekas and Castanon [BeC89], which proposed to adjust an estimate of the cost function of a policy using local corrections along aggregate states that are formed adaptively using the Bellman equation residual. We will discuss later aggregate state formation based on Bellman equation residuals (see Section  3). However, the ideas of the present paper are much broader and aim to find an approximately optimal policy rather than to evaluate exactly a single given policy. Moreover, consistent with the reinforcement learning point of view, our focus in this paper is on large-scale problems with very large number of states $n$.

On the other hand there is a close mathematical connection between classical aggregation ($V=0$) and biased aggregation ($V\ne0$). Indeed, as we will discuss further in the next section, {\it biased aggregation can be viewed as classical aggregation applied to a modified DP problem\/}, which is equivalent to the original DP problem in the sense that it has the same optimal policies. The modified DP problem is obtained from the original by changing its cost per stage from $g(i,u,j)$ to
$$g(i,u,j)-V(i)+\a V(j)),\qquad  i,j\in{\cal S},\ u\in U(i).\xdef\modcost{\lab}\eqnum\show{oneo}$$
Thus the optimal cost function of the modified DP problem, call it $\tl J$, satisfies the corresponding Bellman equation:
$$\tl J(i)=\min_{u\in U(i)}\sum_{j=1}^np_{ij}(u)\big(g(i,u,j)-V(i)+\a V(j)+\a \tl J(j)\big),\qquad  i\in{\cal S},$$
or equivalently
$$\tl J(i)+V(i)=\min_{u\in U(i)}\sum_{j=1}^np_{ij}(u)\Big(g(i,u,j)+\a \big(\tl J(j)+V(j)\big)\Big),\qquad  i\in{\cal S}.\xdef\diffbel{\lab}\eqnum\show{oneo}$$
By comparing this equation with the Bellman Eq.\ \onef\ for the original DP problem, we see that the optimal cost functions of the modified and the original problems are related by
$$\jstar(i)=\tl J(i)+V(i),\qquad  i\in{\cal S},\xdef\vareq{\lab}\eqnum\show{oneo}$$
and that the problems have the same optimal policies. This of course  assumes that the original and modified problems are solved exactly. If instead they are solved approximately using aggregation or another approximation architecture, such as a neural network, the policies obtained may be substantially different. In particular, the choice of $V$ and the approximation architecture may affect substantially the quality of suboptimal policies obtained.

Equation \diffbel\ has been used in various contexts, involving error bounds for value iteration, since the early days of  DP theory, where it is known as the {\it variational form of Bellman's equation\/}, see e.g., [Ber12], Section 2.1.1. The equation can be used to find $\tl J$, which is the variation of $\jstar$ from any guess $V$; cf.\ Eq.\ \vareq. The variational form of Bellman's equation is also implicit in the adaptive aggregation framework of [BeC89]. In reinforcement learning, the variational equation \diffbel\ has been used in various algorithmic contexts under the name {\it reward shaping} or {\it potential-based shaping\/}; see e.g., the papers by Ng, Harada, and Russell [NHR99], Wiewiora [Wie03], Asmuth, Littman, and Zinkov [ALZ08], Devlin and  Kudenko [DeK11], Grzes [Grz17] for some representative works. While reward shaping does not change the optimal policies of the original DP problem, it may change significantly the suboptimal policies produced by approximate DP methods that use linear basis function approximation, such as forms of TD($\l$), SARSA, and others (see [BeT96], [SuB98]). Basically, with reward shaping and a linear approximation architecture,  {\it $V$ is used as an extra basis function\/}. This is closely related with the idea of using approximate cost functions of policies as basis functions, already suggested in the neuro-dynamic programming book [BeT96] (Section 3.1.4).

In the next section we describe the  architecture of our aggregation scheme and the corresponding aggregate problem. In Section 3, we discuss Bellman's equation for the aggregate problem, and the manner in which its solution is affected by the choice of the bias function. We also discuss some of the algorithmic methodology based on the use of the aggregate problem.

\vskip -1pc

\section{Aggregation Framework with a Bias Function}
\vskip -0.5pc

\xdef\figaggregate{\figr}\figrnum\show{myfigure}

\pn The aggregate problem is an infinite horizon Markovian decision problem that involves three sets of states: two copies of the original state space, denoted $I_0$ and $I_1$, as well as a finite set ${\cal A}$ of aggregate states, as depicted in  Fig.\ \figaggregate. The state transitions in the aggregate problem go from a state in ${\cal A}$ to a state in $I_0$, then to a state in $I_1$, and then back to a state in ${\cal A}$, and the process is repeated. At state $i\in I_0$ we choose a control $u\in U(i)$, and then transition to a state $j\in I_1$ at a cost $g(i,u,j)$ according to the original system transition probabilities $p_{ij}(u)$. This is the only type of control in the aggregate problem; the transitions from ${\cal A}$ to $I_0$ and the transitions from $I_1$ to ${\cal A}$ involve no control. Thus policies in the context of the aggregate problem map states $i\in I_0$ to controls $u\in U(i)$, and can be viewed as policies of the original problem. 

The transitions from ${\cal A}$ to $I_0$ and the transitions from $I_1$ to ${\cal A}$ involve two (somewhat arbitrary) choices of transition probabilities, called the {\it
disaggregation and aggregation probabilities\/}, respectively; cf.\  Fig.\ \figaggregate. In particular:

\nitem{(1)} For each aggregate state $x$ and original system state $i$, we specify the {\it
disaggregation probability}
$d_{xi}$, where 
$$\sum_{i=1}^n d_{xi}=1,\qquad \forall\ x\in {\cal A}.$$

\nitem{(2)} For each aggregate state $y$ and original system state $j$, we specify the
{\it aggregation probability}
$\phi_{jy}$, where 
$$\sum_{y\in {\cal A}} \phi_{jy}=1,\qquad \forall\ j\in{\cal S}.$$
\smskip
\pn Several examples of methods to choose the disaggregation and aggregation probabilities are given in Section 6.4 of the book [Ber12], such as hard and soft aggregation, aggregation with representative states, and feature-based aggregation. The latter type of aggregation is explored in detail in the paper [Ber18a], including its use in conjunction with feature construction schemes that involve neural networks and other architectures.

\topinsert{
\centerline{\hskip0pc\includegraphics[width=4.8in]{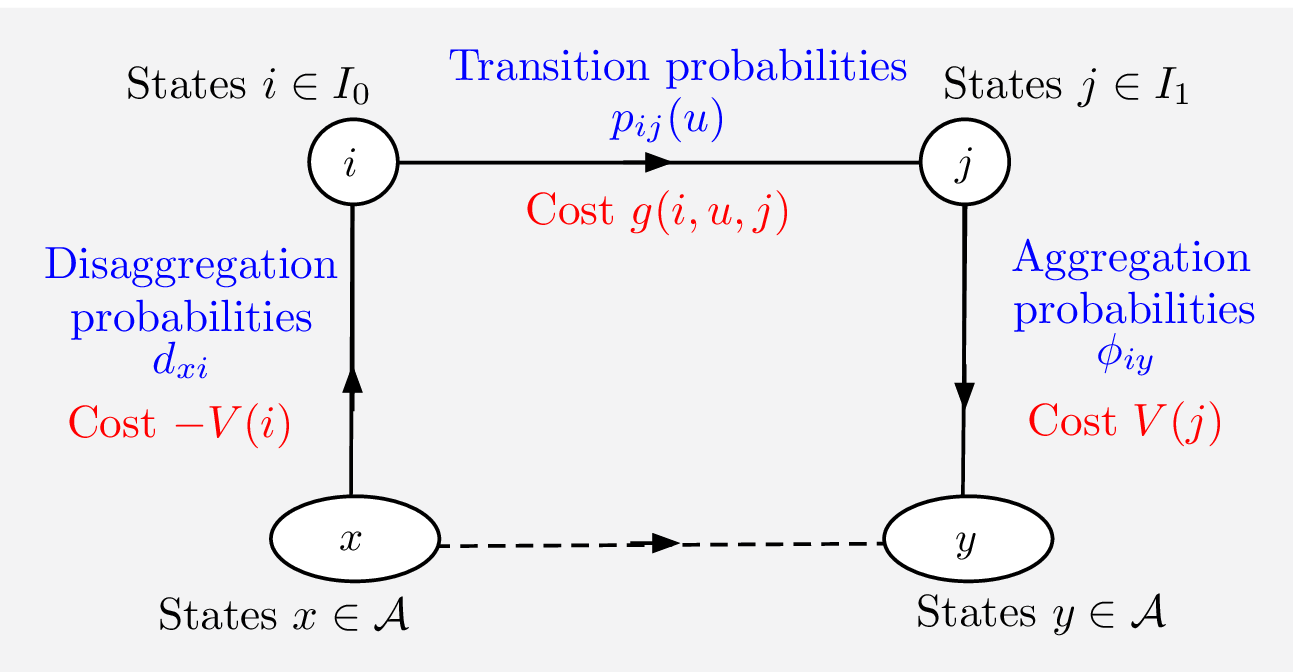}}
\fig{-0.5pc}{\figaggregate}{Illustration of the transition mechanism and the costs per stage of the aggregate problem in the biased aggregation framework. When the bias function $V$ is identically zero, we obtain the classical aggregation framework.\smskip}
}
\endinsert

The definition of the aggregate problem will be complete once we specify the cost for the transitions from ${\cal A}$ to $I_0$ and from $I_1$ to ${\cal A}$. In the classical aggregation scheme, as described in Section 6.4 of [Ber12], these costs are all zero.
The salient new characteristic of the scheme proposed in this paper is a (possibly nonzero) cost $-V(i)$ for transition from any aggregate state to a state $i\in I_0$, and of a cost $V(j)$ from a state $j\in I_1$ to any aggregate state; cf.\  Fig.\ \figaggregate. The function $V$ is called the {\it bias function\/}, and we will argue that {\it $V$ should be chosen as close as possible to $\jstar$\/}. Indeed we will show in the next section that {\it when $V\equiv \jstar$, then an optimal policy for the aggregate problem is also optimal for the original problem\/}, regardless of the choice of aggregate states, and aggregation and disaggregation probabilities. 
Moreover, we will discuss several schemes for choosing $V$, such as for example cost functions of various heuristic policies, in the spirit of the rollout algorithm. 
Generally, while $V$ may be arbitrary, for practical purposes its values at various states should be easily computable. 

It is important to note that the biased aggregation scheme of Fig.\ \figaggregate\ is equivalent to classical aggregation applied to the modified problem where the cost per stage $g(i,u,j)$ is replaced by the cost $g(i,u,j)-V(i)+\a V(j)$ of Eq.\ \modcost. Thus we can straightforwardly transfer results, algorithms, and intuition from classical aggregation to the biased aggregation framework of this paper. In particular, we may use simulation-based algorithms for policy evaluation, policy improvement, and $Q$-learning for the aggregate problem, with the only requirement that the value $V(i)$ for any state $i$ is available when needed.

\vskip-1pc

\section{Bellman's Equation and Algorithms for the Aggregate Problem}
\vskip-0.5pc
\pn The aggregate problem is fully defined as a DP problem once the aggregate states, the aggregation and disaggregation probabilities, and the bias function are specified. Thus its optimal cost function satisfies a Bellman equation, which we will now derive.
To this end, we introduce the cost functions/vectors  
$$\tl r=\big\{\tl r(x)\mid x\in{\cal A}\big\},\qquad \tl J_0=\big\{\tl J_0(i)\mid i\in I_0\big\},\qquad \tl J_1=\big\{\tl J_1(j)\mid j\in I_1\big\},$$
where:

\nitem{} $\tl r(x)$ is the optimal cost-to-go from aggregate state $x$.
\nitem{} $\tl J_0(i)$ is the optimal cost-to-go from state $i\in I_0$ that has just been generated from an aggregate state (left side of Fig.\ \figaggregate).
\nitem{} $\tl J_1(j)$ is the optimal cost-to-go from state $j\in I_1$ that has just been generated from a state $i\in I_0$ (right side of Fig.\ \figaggregate).
\smskip
\pn Note that because of the intermediate transitions to aggregate states, $\tl J_0$ and $\tl J_1$ are different.

These three functions satisfy the following three Bellman equations:
$$\tl r(x)=\sum_{i=1}^nd_{xi}\big(\tl J_0(i)-V(i)\big),\qquad x\in {\cal A},\xdef\belleqo{\lab}\eqnum\show{lsmin}$$
$$\tl J_0(i)=\min_{u\in U(i)}\sum_{j=1}^n p_{ij}(u)\big(g(i,u,j)+\a \tl J_1(j)\big),\qquad i\in{\cal S},\xdef\belleqt{\lab}\eqnum\show{lsmin}$$
$$\tl J_1(j)=V(j)+\sum_{y\in {\cal A}}\phi_{jy}\tl r(y),\qquad j\in{\cal S}.\xdef\belleqth{\lab}\eqnum\show{lsmin}$$
The form of Eq.\ \belleqt\ suggests that {\it $\tl J_0$ and $\tl J_1$ may be viewed as approximations to $\jstar$} (if $\tl J_0$ and $\tl J_1$ were equal they would also be equal to $\jstar$, since Bellman's equation has a unique solution). The form of Eq.\ \belleqo\ suggests that {\it $\tl r$ can be viewed as a ``weighted averaged" variation of $\tl J_0$ from $V$\/}, with weights specified by the disaggregation probabilities $d_{xi}$.

By combining the equations \belleqo-\belleqth, we obtain an equation for $\tl r$:
$$\tl r(x)=(H\tl r)(x),\qquad x\in {\cal A},\xdef\rstfixed{\lab}\eqnum\show{lsmin}$$
where $H$ is the mapping defined by
$$(Hr)(x)=\sum_{i=1}^nd_{xi}\lf(\min_{u\in U(i)}\sum_{j=1}^n p_{ij}(u)\lf(g(i,u,j)+\a\lf(V(j)+ \sum_{y\in {\cal A}}\phi_{jy}r(y)\ri)\ri)-V(i)\ri),\quad x\in {\cal A}.\xdef\aggrmap{\lab}\eqnum\show{lsmin}$$
It can be seen that $H$ is a sup-norm contraction mapping and has $\tl r$ as its unique fixed point. This follows from standard contraction arguments and the fact that $d_{xi}$, $p_{ij}(u)$, and $\phi_{jy}$ are all probabilities (see [Ber12], Section 6.5.2). It also follows from the corresponding result for classical aggregation, by replacing the cost per stage $g(i,u,j)$ with the modified cost 
$$g(i,u,j)+\a V(j)-V(i),\qquad  i,j\in{\cal S},\ u\in U(i),$$ 
of Eq.\ \modcost. 

 In typical applications of aggregation, $\tl r$ has much lower dimension than $\tl J_0$ and $\tl J_1$, and can be found by solving the fixed point equation \rstfixed\ using simulation-based methods, even when the number of states $n$ is very large; this is the major attraction of the aggregation approach.
Once $\tl r$ is found, the optimal cost function $\jstar$ of the original problem may be approximated by the function $\tl J_1$ of Eq.\ \belleqth. Moreover, an optimal policy $\tl \m$ for  the aggregate problem may be found through the minimization in Eq.\ \belleqt\ that defines $\tl J_0$, i.e.,
$$\tl\m(i)\in \arg\min_{u\in U(i)}\sum_{j=1}^n p_{ij}(u)\big(g(i,u,j)+\a \tl J_1(j)\big),\qquad i\in{\cal S}.\xdef\onesteplook{\lab}\eqnum\show{lsmin}$$
The  policy $\tl\m$ is optimal for the original problem if and only if
$\tl J_1$ and $\jstar$ differ uniformly by a constant on the set of states $j$ that are relevant to the minimization above. This is true in particular if $V=\jstar$, as shown by the following proposition.
 
\xdef\propscoreopt{\propn}\propnum\show{myproposition}

\texshopbox{\proposition{\propscoreopt:} When $V=\jstar$ then $\tl r=0$, $\tl J_0=\tl J_1=\jstar$, and any optimal policy for the aggregate problem is optimal for the original problem. }

\proof Since the mapping $H$ of Eq.\ \aggrmap\ is a contraction, it has a unique fixed point. Hence from Eqs.\ \belleqt\ and \belleqth, it follows that the Bellman equations \belleqo-\belleqth\ have as unique solution, the optimal cost functions $(\tl r,\tl J_0,\tl J_1)$ of the aggregate problem.  It can be seen that when $V=\jstar$, then $\tl r=0$, $\tl J_0=\tl J_1=\jstar$ satisfy Bellman's equations  \belleqo-\belleqth, so they are equal to these optimal cost functions. \qed

The preceding proposition suggests that one should aim to choose $V$ as close as possible to $\jstar$. In particular, a good choice of $V$ is one for which the sup-norm of the Bellman equation residual 
 is small, as indicated by the following proposition.

\xdef\propscoreresid{\propn}\propnum\show{myproposition}

\texshopbox{\proposition{\propscoreresid:} We have
$$\|\tl r\|\le {\|V-TV\|\over 1-\a},$$
where $\|\cdot\|$ denotes sup-norm of the corresponding Euclidean space, and $T:\rn\to\rn$ is the Bellman equation mapping 
defined by
$$(TV)(i)= \min_{u\in U(i)}\sum_{j=1}^n p_{ij}(u)\big(g(i,u,j) + \a V(j)\big),
\qquad V\in\rn,\ i\in{\cal S}.$$
}

\proof From Eq.\ \aggrmap\ we have for  any vector $r=\big\{r(x)\mid x\in {\cal A}\big\}$
$$\|Hr\|\le \|TV-V\|+\a \|r\|.$$
Hence by iteration, we have for every $m\ge 1$,
$$\|H^m r\|\le (1+\a+\cdots+\a^{m-1})\|TV-V\|+\a^m\|r\|.$$
By taking the limit as $m\to\infty$ and using the fact $H^m r\to \tl r$, the result follows. \qed

If $V$ is close to $\jstar$, then $\|V-TV\|$ is small (since $\jstar$ is the fixed point of $T$), which implies that $\|\tl r\|$ is small by the preceding proposition. This in turn, by Eq.\ \belleqth, implies that $\tl J_1$ is close to $\jstar$, so that the optimal policy of the aggregate problem  defined by Eq.\ \onesteplook\ is near optimal for the original problem. Choosing $V$ to be close to $\jstar$ may not be easy. A reasonable practical strategy may be to use as $V$ the optimal cost function of a simpler but related problem. Another possibility is to select $V$ to be the cost function of some reasonable policy, or an approximation thereof. We will discuss this possibility in some detail in what follows.

In view of Eq.\ \onesteplook, an optimal policy for the aggregate problem may be viewed as a one-step lookahead policy with lookahead function $\tl J_1$. In the special case where the scalars
$$\sum_{y\in {\cal A}}\phi_{jy}\tl r(y),\qquad  j\in{\cal S},$$
are the same for all $j$, the functions $\tl J_1$ and $V$ differ uniformly by a constant, so an optimal policy for the aggregate problem may be viewed as a one-step lookahead policy with lookahead function $V$. Note that the functions $\tl J_1$ and $V$ differ by a constant in the  extreme case where there is a single aggregate state. If in addition $V$ is equal to the cost function $J_\m$ of a policy $\m$, then an optimal policy for the aggregate problem is a rollout policy based on $\m$ (i.e., the result of a single policy improvement starting with the policy $\m$), as shown by the following proposition. 

\xdef\propscoreroll{\propn}\propnum\show{myproposition}

\texshopbox{\proposition{\propscoreroll:}When there is a single aggregate state and $V=J_\m$ for some policy $\m$, an optimal policy for the aggregate problem is a policy produced by the rollout algorithm based on $\m$. 
}

\proof When there a single aggregate state $y$, we have $\phi_{jy}=1$ for all $j$, so the values $\tl J_1(j)$ and $V(j)$ differ by the constant $\tl r(y)$ for all $j$. Since $V=J_\m$, the optimal policy $\tl \m$ for the aggregate problem is determined by 
$$\tl\m(i)\in \arg\min_{u\in U(i)}\sum_{j=1}^n p_{ij}(u)\big(g(i,u,j)+\a J_\m(j)\big),\qquad i\in{\cal S},\xdef\onesteplook{\lab}\eqnum\show{lsmin}$$
[cf.\ Eq.\ \onesteplook], so $\tl \m$ is a rollout policy based on $\m$. \qed

We next consider the choice of the aggregate states, and its effect on the quality of the solution of the aggregate problem. In particular, biased aggregation produces a correction to the function $V$. The nature of the correction depends on the form of aggregation being used. In the case of hard aggregation, the correction is piecewise constant (it has equal value at all states within each aggregate state). 

We focus on the common case of hard aggregation, but the ideas qualitatively extend to other types of aggregation, such as aggregation with representative states, or feature-based aggregation; see [Ber12], [Ber18a]. For this case the aggregation scheme adds a piecewise constant correction to the function $V$; the correction amount is equal for all states within a given aggregate state. By contrast, aggregation with representative states adds a piecewise linear correction to $V$; the costs of the nonrepresentative states are approximated by linear interpolation of the costs of the representative states, using the aggregation probabilities.

\subsection{Choosing the Aggregate States - Hard Aggregation}
\vskip-0.5pc

\pn  In this section we will  focus on the hard aggregation scheme, which has been discussed extensively in the literature (see, e.g., [BeT96], [TsV96], [Van06], [Ber12], [Ber18a]). The starting point is a partition of the state space ${\cal S}$ that consists of disjoint subsets $I_1,\ldots,I_q$  of states with $I_1\cup\cdots\cup I_q={\cal S}$. The aggregate states are identified with these subsets, so we also use the index $\ell=1,\ldots,q$ to refer to them. The disaggregation probabilities $d_{i\ell}$ can be positive only for states $i\in I_\ell$. To define the aggregation probabilities, let us  denote by $\ell(j)$ the index of the aggregate state to which $j$ belongs. The aggregation probabilities are equal to either 0 or 1, according to aggregate state membership:
$$\phi_{j\ell}=\cases{1&if $\ell=\ell(j)$,\cr
0&otherwise,\cr}\qquad j\in{\cal S},\ \ell=1,\ldots,q.\eqnum\show{qfortyss}$$

In hard aggregation the correction 
$$\sum_{\ell=1}^n\phi_{j\ell}\tl r(\ell),\qquad j\in{\cal S},$$
that is added to $V(j)$ in order to form the aggregate optimal cost function
$$\tl J_1(j)=V(j)+\sum_{\ell=1}^n\phi_{j\ell}\tl r(\ell),\qquad j\in{\cal S},\eqnum\show{lsmin}$$
is piecewise constant: {\it it is constant within each aggregate state $\ell$\/}, and equal to
$$\tl r\big(\ell(j)\big),\qquad j\in{\cal S}.$$
It follows that the success of a hard aggregation scheme depends on both the choice of the bias function $V$ (to capture the rough shape of the optimal cost function $\jstar$) and on the choice of aggregate states (to capture the fine details of the difference $\jstar-V$). This suggests that the variation of $\jstar$ over each aggregate state should be nearly equal to the variation of $V$ over that state, as indicated by the following proposition, which was proved by Tsitsiklis and Van Roy [TsV96] for the classical aggregation case where $V=0$. We have adapted their proof to the biased aggregation context of this paper.

\xdef\properrorbd{\propn}\propnum\show{myproposition}

\texshopbox{\proposition{\properrorbd:}In the case of hard aggregation, where we use a partition of the state space into disjoint sets $I_1,\ldots,I_q$, we have
$$\big|\jstar(i)-V(i)-\tl r(\ell)\big|\le {\e\over 1-\a},\qquad  \forall\ i \hbox{ such that } i\in I_\ell,\ \ell=1,\ldots,q,\eqnum\show{lsmin}$$
where
$$\e=\max_{\ell=1,\ldots,q}\,\max_{i,j\in I_\ell}\big|\jstar(i)-V(i)-\jstar(j)+V(j)\big|.\xdef\epsdef{\lab}\eqnum\show{lsmin}$$
}

\proof Consider the mapping 
$H:\re^q\mapsto\re^q$ defined by
Eq.\ \aggrmap, and consider the vector $\ol r$ with components defined by
$$\ol r(\ell)=\min_{i\in I_\ell}\big(\jstar (i)-V(i)\big)+{\e\over 1-\a},\qquad \ell\in 1,\ldots,q.$$
Using Eq.\ \aggrmap, we have for all $\ell$, we have
$$\eqalignno{(H\ol r)(\ell)&=\sum_{i=1}^n d_{\ell i}\lf(\min_{u\in U(i)}\sum_{j=1}^n p_{ij}(u)\Big(g(i,u,j)+\a \big(V(j)+\ol r\big(\ell(j)\big)\Big)-V(i)\ri)\cr
&\le \sum_{i=1}^nd_{\ell i}\lf(\min_{u\in U(i)}\sum_{j=1}^n p_{ij}(u)\lf(g(i,u,j)+\a \jstar (j)+{\a\e\over 1-\a}\ri)-V(i)\ri)\cr
&=\sum_{i=1}^nd_{\ell i}\lf(\jstar (i)-V(i)+{\a\e\over 1-\a}\ri)\cr
&\le \min_{i\in I_\ell}\big(\jstar(i)-V(i)+\e\br)+{\a\e\over 1-\a}\cr
&=\min_{i\in I_\ell}\big(\jstar(i)-V(i)\big)+{\e\over 1-\a}\cr
&=\ol r(\ell),\cr}$$
where for the second equality we used the Bellman equation for the original system, which is satisfied by $\jstar$, and for the second inequality we used Eq.\ \epsdef. Thus we have $H\ol r\le \ol r$, from which it follows that $\tl r\le \ol r$ (since $H$ is monotone, which implies that the sequence $\{H^k\ol r\}$ is monotonically nonincreasing, and we have 
$$\tl r=\lim_{k\to\infty}H^k\ol r$$
 since $H$ is a contraction and $\tl r$ is its unique fixed point). This proves one side of the desired error bound. The other side follows similarly.
 \qed

The scalar $\e$ of Eq.\ \epsdef\ is the maximum variation of $\jstar-V$ within the sets of the partition of the hard aggregation scheme. Thus the meaning of the preceding proposition is that if $\jstar-V$ changes by at most $\e$ within each set of the partition, the hard aggregation scheme provides a piecewise constant correction $\tl r$ that is within 
${\e/(1-\a)}$ of the optimal $\jstar-V$. 
If the number of aggregate states is sufficiently large so that the variation of $\jstar(i)-V(i)$ within each one is negligible, then the optimal policy of the aggregate problem is very close to optimal for the original problem. 

Finally, let us compare hard aggregation for the classical framework where $V=0$ and for the biased aggregation framework of the present paper where $V\ne 0$. In the former case the optimal cost function $\tl J_1$ of the aggregate problem is piecewise constant (it is constant over each aggregate state). In the latter case $\tl J_1$  consists of $V$ plus a piecewise constant correction (a constant correction over each aggregate state). This suggests that when the number of aggregate states is small, the corrections provided by classical aggregation may be relatively poor,  but the biased aggregation framework may still  perform well with a good choice for $V$. As an example, in the extreme case of a single aggregate state, the classical aggregation scheme results in $\tl J_1$ being constant, which yields a myopic policy, while for $V=J_\m$ the biased aggregation scheme yields a rollout policy based on $\m$. The combined selection of $V$ and the aggregate states so that they work synergistically is an important issue that requires further investigation. 

We will next discuss various aspects of algorithms for constructing a bias function $V$, and for formulating, solving, and using the aggregate problem in various contexts. There are two algorithmic possibilities that one may consider:
\nitem{(a)} Select $V$ using some method that is unrelated to aggregation, and then solve the aggregate problem, obtain its optimal policy, and use it as a suboptimal policy for the original problem. In the case where $V$ is an approximation to the cost function  of some policy, we may view this process as an enhanced form of one-step lookahead. 
\nitem{(b)} Solve the aggregate problem multiple times with different choices of $V$. An example of such a procedure is a policy iteration method, where a sequence of successive policies $\{\m^k\}$ is obtained by solving the aggregate problems with $V\approx J_{\m^k}$.
\smskip
\pn We consider these two possibilities in the next two subsections.

\subsection{An Example of Obtaining a Suboptimal Policy by Biased Aggregation}

\pn In this section we will provide an example that illustrates one possible way to obtain a suboptimal policy by aggregation. We start with a bias function $V$. The nature and properties of $V$ are immaterial for the purposes of this section. Here are some possibilities:

\nitem{(a)} $V$ may be an approximation to $\jstar$ obtained by one or more approximate policy iterations, using a reinforcement learning algorithm such as TD($\l$), LSTD($\l$), or LSPE($\l$), as a linear combination of basis functions (see textbooks such as [BBD10], [BeT96], [Ber12], [Gos15],  [Pow11],  [SuB98],  [Sze10]). Alternatively, $\jstar$ may be approximated using a neural network and simulation-generated cost data, thus obviating the need for knowing suitable basis functions (see e.g., [Ber17]).

\nitem{(b)} $V$ may be an approximation to $\jstar$ obtained in some way that is unrelated to reinforcement learning, such as solving exactly a simplified DP problem. For example, the original problem may be a stochastic shortest path problem, while the simpler problem may be a related deterministic shortest path problem, obtained from the original through some form of certainty equivalence approximation. The deterministic problem can be solved fast to yield $V$ using highly efficient deterministic shortest path methods, even when the number of states $n$ is much larger than the threshold for exact solvability of its stochastic counterpart.
\smskip

Given $V$, we first consider the formation of the aggregate states. The issues are similar to the case of classical hard aggregation, where the main guideline is to select the aggregate states/subsets so that $\jstar$ varies little within each subset. The biased aggregation counterpart of this rule is to select the aggregate states so that $V-\jstar$ varies little within each subset (cf.\ Prop.\ \properrorbd). This suggests that a promising guideline may be to select the aggregate states so that the variation of the $s$-step residual $V-T^sV$ is small within each subset, where $s\ge 1$ is some integer (based on the idea that $T^sV$ is close to $\jstar$). A residual-based approach of this type has been used in the somewhat different context of the paper by Bertsekas and Castanon [BeC89].  Other approaches for selecting the aggregate states based on features are discussed in the survey [Ber18a].  Some further research is needed, both in general and in problem specific contexts, to provide some reliable guidelines for structuring the aggregate problem.

Let us now consider the implementation of the residual-based formation of the aggregate states. We first generate a large sample set of states 
$$\hat{\cal S}=\{i_m\mid m=1,\ldots,M\},$$
and compute the set of their corresponding $s$-step residuals 
$$\big\{V(i)-(T^sV)(i)\mid i\in \hat{\cal S}\big\},\xdef\residset{\lab}\eqnum\show{lsmin}$$
where 
$$(TV)(i)=\min_{u\in U(i)}\sum_{j=1}^np_{ij}(u)\big\{g(i,u,j)+\a V(j)\big\}.$$
We will not discuss here the method of generating the sample set $\hat{\cal S}$. Note that the calculation of $(T^sV)(i)$ requires the solution of an $s$-step DP problem with terminal cost function $V$, and starting state $i$. This may be a substantial calculation, which must of course be carried out off-line. However, it is  facilitated when $s$ is relatively small, the number of controls is relatively small, there are ``few" nonzero probabilities $p_{ij}(u)$ for any $i$ and $u\in U(i)$, and also when parallel computation capability is available. 

We divide the range of the set of residuals \residset\ into disjoint intervals $R_1,\ldots, R_q$, and we group the set of sampled states $\hat{\cal S}$ into disjoint nonempty subsets $I_1,\ldots,I_q$, where $I_\ell$ is the subset of sampled states whose residuals fall within the interval $R_\ell$, $\ell=1,\ldots,q$.\footnote{\dag}
{\ninepoint  More elaborate methods to form the aggregate states are of course possible. A straightforward extension is to partition further the subsets $I_1,\ldots,I_q$ based on some state features, or some other problem-dependent criterion, in the spirit of feature-based aggregation [Ber18a].} The aggregate states are the subsets $I_1,\ldots,I_q$, and the disaggregation probabilities are taken to be equal over each of the aggregate states, i.e.,
$$d_{\ell i}=\cases{1/|I_\ell|& if $i\in I_\ell$,\cr
0&otherwise,\cr}\qquad \ell=1,\ldots,q,\ i\in{\cal S},\xdef\disaggrprob{\lab}\eqnum\show{lsmin}$$
where $|I_\ell|$ denotes the cardinality of the set $I_\ell$ (this is a default choice, there may be other problem-dependent possibilities).
The aggregation probabilities are arbitrary, although it makes sense to require that for $j\in \hat{\cal S}$ we have
$$\phi_{j\ell}=\cases{1& if $j\in I_\ell$,\cr
0&otherwise,\cr}\qquad \ell=1,\ldots,q,\ j\in \hat{\cal S},\xdef\aggrprob{\lab}\eqnum\show{lsmin}$$
and also to choose $\phi_{j\ell}$ based on the degree of ``similarity" of $j$ with states in $I_\ell$ (using for example some notion of ``distance" between states). 

Having specified the aggregate problem, we may solve it by any one of the established simulation-based methods for classical aggregation (see the references noted earlier, and [Ber18a], Section 4.2; see also the next section). These methods apply because the biased aggregation problem corresponding to $V$ is identical to the classical aggregation problem where the cost per stage $g(i,u,j)$ is replaced by 
$g(i,u,j)-V(i)+\a V(j),$
as noted earlier. There are several challenges here, including that the algorithmic solution may be very computation-intensive. On the other hand, algorithms for the aggregate problem aim to solve the fixed point equation \rstfixed, which is typically low-dimensional (its dimension is the number of aggregate states), even when the number of states $n$ is very large. It is also likely that the use of a ``good" bias function will reduce the need for a large number of aggregate states in many problem contexts. This remains to be verified by future research. At the same time we should emphasize that as in classical aggregation, a single policy iteration can produce a policy that is arbitrarily close to optimal, provided the number of aggregate states is sufficiently large. Thus, fewer policy improvements may be needed in aggregation-based policy iteration. 

As an example of a method to solve the aggregate problem, we may consider a stochastic version of the fixed point iteration $r^{t+1}=Hr^t$, where $H:\re^q\mapsto\re^q$ is the mapping with components given by 
$$(Hr)(\ell)=\sum_{i=1}^n d_{\ell i}\lf(\min_{u\in U(i)}\sum_{j=1}^n p_{ij}(u)\lf(g(i,u,j)+\a\lf(V(j)+ \sum_{\ell =1}^q\phi_{j\ell}r(\ell)\ri)\ri)-V(i)\ri),\quad \ell=1,\ldots,q,\xdef\hmapaggr{\lab}\eqnum\show{lsmin}$$
cf.\ Eq.\ \aggrmap.\footnote{\dag}{\ninepoint The other major alternative approach for solving the aggregate problem is simulation-based policy iteration. This algorithm, discussed in [Ber12], Section 6.5.2, and [Ber18a], Section 4.2, generates a sequence of policies $\{\m^k\}$  and corresponding vectors $\{r_{\m^k}\}$ that converge to an optimal policy and cost function $\tl r$ of the aggregate problem, respectively. It involves repeated policy evaluation operations that involve solution of (low-dimensional) fixed point problems of the form $r=H_{\m^k}(r)$, where $H_{\m^k}$ is the mapping given by
$$(H_{\m^k} r)(\ell)=\sum_{i=1}^nd_{\ell i}\sum_{j=1}^n p_{ij}\bl(\m^k(i)\br)\lf(g\bl(i,\m^k(i),j\br)+\a\lf(V(j)+ \sum_{m=1}^q\phi_{jm}\,r(m)\ri)-V(i)\ri),\qquad  \ell=1,\ldots,q,$$
[cf.\ Eq.\ \hmapaggr], interleaved policy improvement operations that define $\m^{k+1}$ from the fixed point $r_{\m^k}$ of $H_{\m^k}$ by
$$\m^{k+1}(i)=\arg\min_{u\in U(i)}\sum_{j=1}^n p_{ij}(u)\lf(g(i,u,j)+\a\lf(V(j)+  \sum_{\ell=1}^q\phi_{j\ell}r_{\m^k}(\ell)\ri)\ri),\qquad i=1,\ldots,n.$$
The fixed point problems involved in the policy evaluations are linear, so they can be solved using simulation-based algorithms similar to TD($\l$), LSTD($\l$), and LSPE($\l$) (see e.g., [Ber12] and other reinforcement learning books). For detailed coverage of simulation-based methods for solving general linear systems of equations, see the papers by Bertsekas and Yu [BeY07], [BeY09], Wang and Bertsekas [WaB13a], [WaB13b], and the book [Ber12], Section 7.3.
}
 This algorithm, due to Tsitsiklis and Van Roy [TsV96],  generates a sequence of aggregate states 
 $$\{I_{\ell_t}\mid t=0,1,\ldots\}$$
  by some probabilistic mechanism, which ensures that all aggregate states are generated infinitely often. Given $r^t$ and $I_{\ell_t}$, the algorithm generates an original system state $i_t\in I_{\ell_t}$ according to the uniform probabilities $d_{\ell i}$, and updates the component $r({\ell_t})$ according to
$$r^{t+1}({\ell_t})=(1-\g_t)r^t({\ell_t})+\g_t \lf(\min_{u\in U(i)}\sum_{j=1}^n p_{i_t j}(u)\lf(g(i_t,u,j)+\a \lf(V(j)+\sum_{\ell =1}^q\phi_{j\ell}r^t_{\ell}\ri)-V({i_t})\ri)\ri),\xdef\stochvi{\lab}\eqnum\show{lsmin}$$
where $\g_t$ is a positive stepsize, and leaves all the other components unchanged:
$$r^{t+1}(\ell)=r^t(\ell),\qquad \hbox{if }\ell\ne \ell_t.$$
The stepsize $\g_t$ should be diminishing (typically at the rate of $1/t$). We refer to the paper [TsV96] for further discussion and analysis (see also [BeT96], Section 3.1.2 and 6.7).

Under appropriate mild assumptions, the iterative algorithm \stochvi\ is guaranteed to yield in the limit the optimal cost function $\tl r=\big(\tl r(1),\ldots,\tl r(q)\big)$ of the aggregate problem, where $\tl r(\ell)$ corresponds to aggregate state $I_\ell$. Once this happens, the optimal policy of the aggregate problem is obtained from the minimization
$$\tl\m(i)\in\arg\min_{u\in U(i)}\sum_{j=1}^n p_{ij}(u)\lf(g(i,u,j)+\a\lf(V(j)+ \sum_{\ell =1}^q\phi_{j\ell}\tl r(\ell)\ri)\ri),\qquad i\in{\cal S}.\xdef\imprpol{\lab}\eqnum\show{lsmin}$$

We note that the convergence of the iterative algorithm \stochvi\ can be substantially enhanced if a good initial condition $r^0$ is known. Such an initial condition can be obtained in the special case where $U(i)$ is the same and equal to some set $U_\ell$ for all $i\in I_\ell$. Then $r^0$ can be set to $\hat r=\big(\hat r(1),\ldots,\hat r(q)\big)$, the optimal cost function of the simpler aggregate problem, where the policy is restricted to use the same control for all $i\in I_\ell$ (see [Ber12], Section 6.5.1). One way to obtain $\hat r$ is by solving the low-dimensional linear programming problem of maximizing $\sum_{\ell=1}^q \hat r(\ell)$ subject to the constraint $\hat r\le H(\hat r)$, or 
\maxprob{\sum_{\ell=1}^q \hat r(\ell)}{\hat r(\ell)\le \sum_{i\in I_\ell}d_{\ell i}\sum_{j=1}^n p_{i j}(u)\lf(g(i,u,j)+\a \lf(V(j)+\sum_{s =1}^q\phi_{js}\hat r(s)\ri)-V({i})\ri),\quad \ell =1,\ldots,q, \ u\in U_\ell;}{0} 
see [Ber12], Sections 2.4 and 6.5.1.

By its nature, the stochastic iterative algorithm \stochvi\ must be implemented off-line; this is also true for other solution methods for solving the aggregate problem, such as simulation-based policy iteration. On the other hand, after the limit $\tl r$ is obtained, the improved policy $\tl \m$ of Eq.\ \imprpol\ cannot be computed and stored off-line when the number of states $n$ is large, so it must be implemented on-line.
This requires forming the expectation in Eq.\ \imprpol\ for each $u\in U(i)$, which can be prohibitively time-consuming. An alternative is to calculate off-line approximate $Q$-factors
$$\tl Q(i,u)\approx \sum_{j=1}^n p_{ij}(u)\lf(g(i,u,j)+\a\lf(V(j)+ \sum_{\ell =1}^q\phi_{j\ell}\tl r(\ell)\ri)\ri),\qquad i\in{\cal S},\old{\eqnum\show{lsmin}}$$
using $Q$-factor samples and a neural network or other approximation architecture. One can then implement the  policy $\tl \m$ by means of the simpler calculation
$$\tl\m(i)\in\arg\min_{u\in U(i)}\tl Q(i,u),\qquad i\in{\cal S},$$
which does not require forming an expectation.

Let us now compare the preceding algorithm with the rollout algorithm, which is just Eq.\ \imprpol\ with $V$ equal to the cost function of a base policy and $\tl r=0$. The main difference is that the rollout algorithm bypasses the solution of the aggregate problem, and thus avoids the off-line calculation of $\tl r$. Thus a single policy improvement using the preceding aggregation-based algorithm may be viewed as an enhanced form of rollout, where a better performing policy is obtained at the expense of substantial off-line computation.

\subsection{Aggregation-Based Approximate Policy iteration}

\pn In this section we  consider an approximate policy iteration-like scheme, which  each iteration $k$ starts with a policy $\m^k$, and produces an optimal policy $\m^{k+1}$ of the aggregate problem corresponding to $V_k\approx J_{\m^{k}}$. We discuss  general aspects of this process here, and in the next subsection we provide an example implementation, where $J_{\m^{k}}$ is itself approximated by aggregation. There are three main issues:

\nitem{(a)} How to calculate bias function values $V_k(i)$ that are good approximations to $J_{\m^{k}}(i)$.
\nitem{(b)} How to complete the definition of the aggregate problem, i.e., select the aggregate states, and the aggregation and disaggregation probabilities.
\nitem{(c)} How to obtain $\m^{k+1}$ as an approximately optimal solution of the aggregate problem corresponding to $V_k$, the aggregate states, and the aggregation/disaggregation probabilities defined by (a) and (b) above.
\smskip

Regarding question (a), one possibility is to use Monte-Carlo simulation to approximate $J_{\m^k}(i)$ for any state $i$ as needed, in the spirit of the rollout algorithm. Another possibility is to introduce a  simulation-based policy evaluation to approximate $J_{\m^k}$. For example, we may use TD($\l$), LSTD($\l$), or LSPE($\l$), or a neural network and simulation-generated cost data. Note that such a policy evaluation phase is separate and must be conducted before starting the biased aggregation-based policy improvement that produces the policy $\m^{k+1}$. A related approach is to aim for a bias function that is closer to $\jstar$ than $J_{\m^k}$ is. This may be attempted by using a neural network-based approach based on optimistic policy iteration or $Q$-learning method such as SARSA and its variants; see [SuB98] and other reinforcement learning textbooks cited earlier.

Regarding question (b), the issues are similar to the situation discussed in the preceding section. As noted there, a promising guideline is to select the aggregate states so that the variation of the $s$-step residual $V_k-T^sV_k$ is small within each, where $s\ge 1$ is some integer. Note that the calculation of the residuals is simpler than in the case of the preceding section because in the context of the present section only one policy is involved. 
Similarly, regarding question (c), one may use any of the established simulation-based methods for classical aggregation, as noted in the preceding section.

In the special case where we start with some policy and perform just a single policy iteration, the method may be viewed as an enhanced version of the rollout algorithm. Such a rudimentary form of policy iteration could be the method of choice in a given problem context, because the ``improved" policy may be implemented on-line by simple Monte Carlo simulation, similar to the rollout algorithm. This is particularly so in deterministic problems, such as scheduling, routing, and other combinatorial optimization settings, where the rollout algorithm has been used with success.

\subsubsection{Error Bound for Approximate Policy Improvement}

\pn Let us now provide an error bound for the difference $J_{\tl \m}-J_{\m}$, where $\m$ is a given policy, and $\tl \m$ is an optimal policy for the aggregate problem with $V=J_\m$. 
By Prop.\ \propscoreopt, the Bellman equations for $\m$ have the form
$$r_\m(x)=\sum_{i=1}^nd_{xi}\big(J_{0,\m}(i)-J_\m (i)\big),\qquad x\in {\cal A},\eqnum\show{lsmin}$$
$$J_{0,\m}(i)=\sum_{j=1}^n p_{ij}\big(\m(i)\big)\Big(g\big(i,\m(i),j\big)+\a J_{1,\m}(j)\Big),\qquad i\in{\cal S},\eqnum\show{lsmin}$$
$$J_{1,\m}(j)=J_\m (j)+\sum_{y\in {\cal A}}\phi_{jy}r_\m (y),\qquad j\in{\cal S}.\eqnum\show{lsmin}$$
and their unique solution, the cost function of $\m$ in the aggregate problem, is $(r_\m,J_{0,\m},J_{1,\m})$ given by 
$$r_\m=0,\qquad J_{0,\m}=J_{1,\m}=J_\m.$$
It follows that the optimal cost function $(\tl r,\tl J_0,\tl J_1)$ of the aggregate problem satisfies
$$\tl r\le r_\m=0,\qquad \tl J_0\le J_{0,\m}=J_\m,\qquad \tl J_1\le J_{1,\m}=J_\m.\xdef\aggrreduced{\lab}\eqnum\show{lsmin}$$
Moreover  optimal policy $\tl \m$ for the aggregate problem satisfies
$$\tl\m(i)\in \arg\min_{u\in U(i)}\sum_{j=1}^n p_{ij}(u)\big(g(i,u,j)+\a \tl J_{1}(j)\big),\qquad i\in{\cal S}.\eqnum\show{lsmin}$$

From Eq.\ \belleqth, we have
$$\tl J_1(j)=J_\m(j)+\sum_{y\in {\cal A}}\phi_{jy}\tl r(y)),\qquad i\in{\cal S}.$$
Using the fact $\tl r\le 0$ in the preceding equation, we obtain $\tl J_1\le J_\m$, so that
$$T_{\tl \m}\tl J_1=T\tl J_1\le T J_\m\le T_\m J_\m=J_\m,$$
where we have introduced the Bellman equation mappings $T_\m:\rn\to\rn$, given by
$$(T_\m J)(i)= \sum_{j=1}^n p_{ij}\big(\m(i)\big)\big(g\big(i,\m(i),j\big)+\a J(j)\big),\qquad J\in\rn,\ i\in{\cal S},$$
and  $T_{\tl \m}:\rn\to\rn$, given by
$$(T_{\tl \m} J)(i)= \sum_{j=1}^n p_{ij}\big({\tl \m}(i)\big)\big(g\big(i,{\tl \m}(i),j\big)+\a J(j)\big),\qquad J\in\rn,\ i\in{\cal S}.$$

By combining the preceding two relations, we obtain
$$(T_{\tl \m}J_\m)(i)+\g(i)\le J_\m(i),\qquad i\in{\cal S},\xdef\epsapprox{\lab}\eqnum\show{lsmin}$$
where $\g(i)$ is the number 
$$\g(i)=\a\sum_{j=1}^np_{ij}\big(\tl \m(i)\big)\sum_{y\in {\cal A}}\phi_{jy}\tl r(y),$$
which is nonpositive since $\tl r\le0$ [cf.\ Eq.\ \aggrreduced].
From Eq.\ \epsapprox\ it follows, by repeatedly applying $T_{\tl \m}$ to both sides,  that for all $m\ge 1$, we have 
$$(T^m_{\tl \m}J_\m)(i)+(\g+\a\g+\cdots+\a^{m-1}\g)\le J_\m(i),\qquad i\in{\cal S},$$
where
$$\g=\min_{i\in{\cal S}}\g(i),$$
so that by taking the limit as $m\to\infty$ and using the fact $T^m_{\tl \m}J_\m\to J_{\tl\m}$, we obtain
$$J_{\tl\m}(i)\le J_{\m}(i)-{\g\over 1-\a},\qquad i\in{\cal S}.$$
Thus solving the aggregate problem with $V$ equal to the cost function of a policy yields only an approximate policy improvement for the original problem, with an approximation error that is bounded by $-{\g\over 1-\a}$.

Let us return now to the policy iteration-like scheme that produces at iteration $k$ an optimal policy $\m^{k+1}$ of the aggregate problem corresponding to $V_{k}=J_{\m^{k}}$. From the preceding analysis it follows that even if $J_{\m^{k}}$ is computed exactly, this scheme will not produce an optimal policy of the original problem. Instead, similar to typical approximate policy iteration schemes, the policy cost functions $J_{\m^k}$ will oscillate within an error bound that depends on the aggregation structure (the aggregate states, and the aggregation and aggregation probabilities); see [BeT96], Section 6.2.2. This suggests that there is potential benefit for changing the aggregation structure with each iteration, as well as for not solving each aggregate problem to completion (as in ``optimistic" policy iteration; see [BeT96]). Further research may shed some light into these issues.

\subsection{An Example of Policy Evaluation by Aggregation}
\pn We will now consider the use of biased aggregation to evaluate approximately the cost function of a fixed given policy $\m$, thus providing an alternative to Monte Carlo simulation as in rollout, or basis function approximation methods such as TD($\l$), LSTD($\l$), or LSPE($\l$), or neural network-based policy evaluation. We start with some initial function $\hat J_0$, and we generate a sequence of functions $\{\hat J_k\}$ that converges to an approximation of $J_\m$. The idea is to mix approximate value iterations for the original problem, with  low-dimensional aggregate value iterations, with the aim of approximating $J_\m$, while bypassing high-dimensional calculations of order $n$. This is inspired by the iterative aggregation ideas of Chatelin and Miranker [ChM82] for solving linear systems of equations, and the adaptive aggregation ideas of Bertsekas and Castanon [BeC89], but differs in one important respect: in both papers [ChM82] and [BEC89], the aim is to compute $J_\m$ exactly, so the value iterations should be exact and should involve all states $1,\ldots,n$; this restricts applicability to problems where the value of $n$ is modest. By contrast, in our framework the value iterations involve only a subset of the states, which makes our approach applicable to large-scale problems. The price for this is that we cannot hope to obtain $J_\m$ exactly. We can only aspire to a ``good" approximation of $J_\m$.

The methodology of the paper [ChM82] is also different in that the aggregation framework is static and does not change from one iteration to the next. By contrast, similar to the paper [BeC89], the aggregation framework of this section is adaptive and is changed at the start of each aggregation iteration, with the aggregate states formed based on magnitude of Bellman equation residuals.  

Since we will be dealing with a single policy $\m$, to simplify notation, we will abbreviate $p_{ij}\big(\m(i)\big)$, $g\big(i,\m(i),j\big)$, and $T_\m$,   with $p_{ij}$, $g(i,j)$, and $T$, respectively. Thus for the purposes of this section, we use the notation
$$(TJ)(i)=\sum_{j=1}^n p_{ij}\big(g(i,j)+\a J(j)\big),\qquad i\in{\cal S},\ J\in\rn.\xdef\maptmu{\lab}\eqnum\show{lsmin}$$
We introduce a large sample set of states 
$$\hat{\cal S}=\{i_m\mid m=1,\ldots,M\},$$ which may remain fixed through the algorithm, or may be redefined at the beginning of each iteration.

At the beginning of iteration $k$, we have a function $\hat J_k=\big(\hat J_k(1),\ldots,\hat J_k(n)\big)$. We do not assume that this function is stored in memory, since we do not want to preclude situations where $n$ is very large. Instead we assume that $\hat J_k(i)$ can be calculated for any given state $i$, when needed. 
Similar to Subsection 3.2, we generate 
and compute the multistep residuals for the sample states
$$\big\{\hat J_k(i)-(T^{s_k}\hat J_k)(i)\mid i\in\hat{\cal S}\big\},\xdef\multires{\lab}\eqnum\show{lsmin}$$
where $s_k\ge 1$ is an integer. As noted earlier, this is a feasible calculation using the expression \maptmu, even for a large-scale problem, provided the transition probability matrix of $\m$ is sparse (there are ``few" nonzero probabilities $p_{ij}$ for any $i$). The reason is that $(T^{s_k}\hat J_k)(i)$ can be calculated with an $s_k$-step DP calculation as the cost accumulated by the policy $\m$ over $s_k$ steps with terminal cost function $\hat J_k$ and starting from $i$. In particular, to calculate $(T^{s_k}\hat J_k)(i)$ we need to generate the tree of $s_k$-step transition paths starting from $i$, and accumulate the costs along these paths [including $\hat J_k(j)$ at the leaf states $j$ of the tree], weighted by the corresponding probabilities (this is why we need the values of $\hat J_k$ at all states $j$, not just on the states in $\hat{\cal S}$).

Next, as earlier, we divide the range of the residuals \multires\ into disjoint intervals $R_1,\ldots, R_q$, and we group the set of sampled states $\hat{\cal S}$ into disjoint nonempty subsets $I_1,\ldots,I_q$, where $I_\ell$ is the subset of  states whose residuals fall within the interval $R_\ell$, $\ell=1,\ldots,q$. The aggregate states are the subsets $I_1,\ldots,I_q$, and the disaggregation and aggregation probabilities are formed similar to the preceding subsection [cf.\ Eqs.\ \disaggrprob\ and \aggrprob].\footnote{\dag}{\ninepoint There are a few questions left unanswered here, such as the method to generate the sample states, the selection of the number of value iterations $s_k$, the selection of aggregation/disaggregation probabilities, etc. Also there are several variants of the method for forming the aggregate states. For example, the aggregate states may be grouped based on the values of the single step residuals
$$\big\{(T^{s_k-1}\skew5\hat J_k)(i)-(T^{s_k}\skew5\hat J_k)(i)\mid i\in\skew5\hat{\cal S}\big\},$$
rather than the multistep residuals \multires, and they may be further subdivided based on some state features or some problem-dependent criterion.
We leave these questions aside for the moment, recognizing that to address them requires experimentation in a variety of problem-dependent contexts.} 

To complete the aggregation framework, we specify the bias function to be 
$$V_k(i)=(T^{s_k-1}\skew5\hat J_k)(i),\qquad i\in \hat{\cal S},$$
so that
$$TV_k(i)=(T^{s_k}\skew5\hat J_k)(i),\qquad i\in \hat{\cal S}.$$
We set $\hat J_{k+1}$ to be the aggregate cost function obtained from the aggregation framework just specified. In particular, we first solve the aggregate problem, and obtain the vector $\hat r_k$, the fixed point of the corresponding mapping $H$, cf.\ Eq.\ \aggrmap. This equation in the context of the present section takes the form
$$\eqalign{(Hr)(\ell)&=\sum_{i=1}^nd_{\ell i}\lf((TV_k)(i)-V_k(i)+\a\sum_{j=1}^n p_{ij} \sum_{\ell=1}^q\phi_{j\ell}r(\ell)\ri)\cr
&=\sum_{i=1}^nd_{\ell i}\lf((T^{s_k}\skew5\hat J_k)(i)-(T^{s_k-1}\skew5\hat J_k)(i)+\a\sum_{j=1}^n p_{ij} \sum_{\ell=1}^q\phi_{j\ell}r(\ell)\ri),\qquad \ell=1,\ldots,q,\cr}$$
where the first equality follows from Eq.\ Eq.\ \aggrmap, and the fact that we are dealing with a single policy, so there is no minimization over $u$. We do this with either the iterative method \stochvi\ (without the minimization over $u$, since we are dealing with a single policy), or by matrix inversion that computes the fixed point of the mapping $H$ (which is linear and low-dimensional). 
We then define the function $\hat J_{k+1}$ on the sample set of states by
$$\hat J_{k+1}(i)=(T^{s_k}\skew5\hat J_k)(i)+\a\sum_{j=1}^np_{ij}\sum_{\ell=1}^q\phi_{j\ell}\hat r_k(\ell),\qquad i\in \hat{\cal S}.$$
Note that $\hat J_{k+1}$ is the result of $s_k$ value iterations applied to $\hat J_k$, followed by a correction determined from the solution to the aggregate problem. Proposition \propscoreresid\ suggests that it is desirable that  $\|V_k-TV_k\|$ is small. This in turns indicates that $s_k$ should be chosen sufficiently large, to the point where the value iterations are converging slowly.

The final step before proceeding to the next iteration is to extend the definition of $\hat J_{k+1}$ from $\hat{\cal S}$ to the entire state space ${\cal S}$. One possibility for doing this is through a form of interpolation using some nonnegative weights 
$$\xi_{ji},\qquad j\in{\cal S},\ i\in \hat{\cal S},$$
with 
$$\sum_{i\in \hat{\cal S}}\xi_{ji}=1,\qquad j\in{\cal S},\xdef\xinormal{\lab}\eqnum\show{qfortyss}$$
and to define 
$$\hat J_{k+1}(j)=\sum_{i\in \hat{\cal S}}\xi_{ji}\hat J_{k+1}(i),\qquad j\in{\cal S}.$$
A possible choice is to use the weights
$$\xi_{ji}=\sum_{\ell=1}^q\phi_{j\ell}d_{\ell i},\qquad j\in{\cal S},\ i\in \hat{\cal S},\xdef\interweights{\lab}\eqnum\show{qfortyss}$$
which intuitively makes sense and satisfies the normalization condition \xinormal\ since
$$\sum_{i\in \hat{\cal S}}\xi_{ji}=\sum_{i\in \hat{\cal S}}\sum_{\ell=1}^q\phi_{j\ell}d_{\ell i}=
\sum_{\ell=1}^q\phi_{j\ell}\sum_{i\in \hat{\cal S}}d_{\ell i}=\sum_{\ell=1}^q\phi_{j\ell}=1,\qquad  j\in{\cal S}.$$
With this last step the definition $\hat J_{k+1}(i)$ for all states $i$ is complete, and we can proceed to the next iteration. In the case where the sampled set $\hat {\cal S}$ is equal to the entire state space ${\cal S}$, this final step is unnecessary. Then the algorithm becomes very similar to the one of the paper [BeC89]. As discussed in that paper, with appropriate safeguards, the sequence $\{\hat J_k\}$ is guaranteed to converge to $J_\m$ thanks to the convergence property of the value iteration algorithm.
 
Let us finally note that the ideas of this section are applicable and can be extended to the approximate solution of general  contractive linear systems of equations, possibly involving infinite dimensional continuous-space operators. The aggregation-based algorithm can be viewed as a hierarchical up-and-down sampling process. Starting with an approximate solution $\hat J_k$ defined on the original state space ${\cal S}$, we downsample to a lower-dimensional state space defined by the sampled set of states $\hat {\cal S}$. We divide this set into aggregate states/subsets $I_1,\ldots,I_q$, and formulate an aggregate problem whose solution 
$$\hat r_k=\big(\hat r_1(1),\ldots,\hat r_k(q)\big),$$
defines a function $\hat J_{k+1}$ on the set $\hat {\cal S}$. Finally, $\hat J_{k+1}$ is upsampled to the original state space using linear interpolation weights such as those of Eq.\ \interweights.


\vskip-1.5pc

\section{Concluding Remarks}
\vskip-0.5pc
\pn In this paper we have proposed  a new aggregation framework, which provides a connection with several successful reinforcement learning approaches, such as rollout algorithms, approximate policy iteration, and other single and multistep lookahead methods. The key is the use of a {\it bias function\/}, which biases the values of the aggregate cost function towards their correct levels. 

An important issue within our aggregation context is the choice of the bias function $V$. In this paper, we have paid some attention to the choice $V=J_\m$ for some base policy $\m$, which highlighted the connection with rollout and approximate policy iteration algorithms. On the other hand, $V$ can be any reasonable approximation to $\jstar$, however obtained, including through the use of simulation-based approximation in value space, and neural networks or other approximation architectures.  Another interesting related issue is the use of multiple bias functions that may be linearly combined with tunable weights to form a single bias function $V$. Generally, the choice of $V$, the formation of the corresponding biased aggregation framework, and attendant computational experimentation are subjects that require further research.

In this paper, we have focused on discounted problems, but our approach applies to all the major types of DP problems, including finite horizon, discounted, and stochastic shortest path problems. Of special interest are deterministic discrete-state problems, which arise in combinatorial optimization. For such problems, rollout algorithms have been used with success, and have provided substantial improvements over the heuristics on which they are based. One may try to improve the rollout algorithms for these problems with the use of biased aggregation. 

We finally note that aggregation can be implemented in several different contexts, such as multistage or distributed aggregation (see Sections 6.5.3, 6.5.4, and [Ber18b], Section 1.2). The idea of introducing a bias function within these contexts in ways similar to the one of the present paper is straightforward, and is an interesting subject for further investigation.

\vskip-1.5pc
\section{References}
\vskip-0.5pc

\def\ref{\vskip1.pt\pn}

\ref[ALZ08] Asmuth, J., Littman, M.\ L. and Zinkov, R., 2008.\ ``Potential-Based Shaping in Model-Based Reinforcement Learning," Proc.\ of 23rd AAAI Conference, pp.\ 604-609.

\ref[BBD10] Busoniu, L., Babuska, R., De Schutter, B., and Ernst, D.,  2010.\ Reinforcement Learning and Dynamic Programming Using Function Approximators, CRC Press, N.\ Y.

\ref[BBS87] Bean, J.\ C., Birge, J.\ R., and Smith, R.\ L., 1987.\ ``Aggregation in Dynamic Programming," Operations Research, Vol.\ 35, pp.\ 215-220.

\ref [BeC89] Bertsekas, D.\ P., and Castanon, D.\ A., 1989.\ 
``Adaptive  Aggregation Methods for Infinite Horizon Dynamic Programming," IEEE
Trans.\ on Aut.\  Control, Vol.\ AC-34, pp.\ 589-598.

\ref [BeT91]  Bertsekas, D.\ P., and Tsitsiklis, J.\ N., 1991.\ ``An Analysis of
Stochastic Shortest Path Problems,"
Math.\ Operations Research, Vol.\ 16, pp.\ 580-595.

\ref [BeT96]  Bertsekas, D.\ P., and Tsitsiklis, J.\ N., 1996.\ Neuro-Dynamic
Programming, Athena Scientific, Belmont, MA.

\ref [BeY07]  Bertsekas, D.\ P., and Yu, H., 2007.\ ``Solution of Large Systems of Equations Using 
Approximate Dynamic Programming Methods," Lab.\
for Information and Decision Systems Report LIDS-P-2754, MIT.

\ref [BeY09]  Bertsekas, D.\ P., and Yu, H., 2009.\ ``Projected Equation Methods for Approximate Solution of Large Linear Systems," J.\ of Computational and Applied Mathematics, Vol.\ 227, pp.\ 27-50.

\ref[Ber12] Bertsekas, D.\ P., 2012.\ Dynamic Programming and Optimal Control, Vol.\ II, 4th edition, Athena Scientific, Belmont, MA.

\ref[Ber17] Bertsekas, D.\ P., 2017.\ Dynamic Programming and Optimal Control, Vol.\ I, 4th edition, Athena Scientific, Belmont, MA.

\ref[Ber18a] Bertsekas, D.\ P., 2018.\ ``Feature-Based Aggregation and Deep Reinforcement Learning: A Survey and Some New Implementations," Lab. for Information and Decision Systems Report, MIT, April 2018 (revised August 2018); arXiv preprint arXiv:1804.04577; will appear in IEEE/CAA Journal of Automatica Sinica. 

\ref[Ber18b] Bertsekas, D.\ P., 2018.\ Abstract Dynamic Programming, Athena Scientific, Belmont, MA.

\ref[Ber19] Bertsekas, D.\ P., 2019.\ Reinforcement Learning and Optimal Control, Athena Scientific, Belmont, MA.

\ref[ChM82] Chatelin, F., and Miranker, W.\ L., 1982.\  ``Acceleration by Aggregation of Successive Approximation Methods," Linear Algebra and its Applications, Vol.\ 43, pp.\ 17-47.

\ref[CiS15] Ciosek, K., and Silver, D., 2015.\ ``Value Iteration with Options and State Aggregation," Report, Centre for Computational Statistics and Machine Learning University College London.

\ref[DeK11] Devlin, S.,  and Kudenko, D., 2011.\ ``Theoretical Considerations of Potential-Based Reward Shaping for Multi-Agent Systems," In Proceedings of AAMAS.

\ref[Gor95] Gordon, G.\ J., 1995.\ ``Stable Function Approximation in  Dynamic Programming,''
in Machine Learning: Proceedings of the 12th International Conference, Morgan Kaufmann, San Francisco, CA.

\ref[Gos15] Gosavi, A., 2015.\ Simulation-Based Optimization:
Parametric Optimization Techniques and Reinforcement Learning, 2nd Edition, Springer, N.\ Y.

\ref[Grz17] Grzes, M., 2017.\ ``Reward Shaping in Episodic Reinforcement Learning," in Proc.\ of the 16th Conference on Autonomous Agents and MultiAgent Systems, pp.\ 565-573.

\ref[NHR99] Ng, A.\ Y., Harada, D., and Russell, S.\ J., 1999.\  ``Policy Invariance Under Reward Transformations: Theory and Application to Reward Shaping," in Proc.\ of the 16th International Conference on Machine Learning, pp.\ 278-287.

\ref [Pow11] Powell, W.\ B., 2011.\  Approximate Dynamic Programming: Solving the Curses of Dimensionality, 2nd Edition, J.\ Wiley and Sons, Hoboken, N.\ J.

\ref[RPW91] Rogers, D.\ F., Plante, R.\ D., Wong, R.\ T.,  and Evans, J.\ R., 1991.\ ``Aggregation and Disaggregation Techniques and Methodology in Optimization," Operations Research, 
Vol.\ 39, pp.\ 553-582.

\ref[SJJ95] Singh, S.\ P., Jaakkola, T., and Jordan, M.\ I., 1995.\
``Reinforcement Learning with Soft State Aggregation,'' in Advances in Neural
Information Processing Systems 7,  MIT Press, Cambridge, MA.

\ref[SSP18] Serban, I.\ V., Sankar, C., Pieper, M., Pineau, J., Bengio, J., 2018.\ ``The Bottleneck Simulator:
A Model-Based Deep Reinforcement Learning Approach," arXiv preprint arXiv:1807.04723.v1. 

\ref[SuB98] Sutton, R.\  S., and Barto, A.\ G., 1998.\ Reinforcement Learning, MIT
Press, Cambridge, MA. (A draft 2nd edition is available on-line.)

\ref [Sze10] Szepesvari, C., 2010.\ Algorithms for Reinforcement Learning, Morgan and Claypool Publishers, San Franscisco, CA.

\ref[TsV96] Tsitsiklis, J.\ N., and Van Roy, B., 1996. ``Feature-Based Methods for
Large-Scale Dynamic Programming," Machine Learning, Vol.\ 22, pp.\ 59-94.

\ref [VDR79] Vakhutinsky, I. Y., Dudkin, L.\ M., and Ryvkin, A.\ A., 1979.\ ``Iterative Aggregation - A New Approach to the Solution of Large Scale Problems," Econometrica, Vol.\ 47, pp.\ 821-841.

\ref[Van06] Van Roy, B., 2006.\ ``Performance Loss Bounds for Approximate Value Iteration with State Aggregation," Mathematics of Operations Research, Vol.\ 31, pp.\ 234-244.

\ref[WaB13a] Wang, M.,  and Bertsekas, D.\ P., 2013.\ ``Stabilization of Stochastic Iterative Methods for Singular and Nearly Singular Linear Systems," Mathematics of Operations Research, Vol.\ 39, pp.\ 1-30.

\ref[WaB13b] Wang, M.,  and Bertsekas, D.\ P., 2013.\ ``Convergence of Iterative Simulation-Based Methods for Singular Linear Systems," Stochastic Systems, Vol.\ 3, pp.\ 39-96.

\ref[Wie03] Wiewiora, E., 2003.\ Potential-Based Shaping and Q-Value Initialization are Equivalent," J.\ of Artificial Intelligence Research, Vol.\ 19, pp.\ 205-208.

\ref[YuB04] Yu, H., and Bertsekas, D.\ P., 2004.\ ``Discretized Approximations for POMDP with Average Cost," Proc.\ of the 20th Conference
on Uncertainty in Artificial Intelligence, Banff, Canada.

\end